\documentclass[10pt,twocolumn,letterpaper]{article}
 
\usepackage{booktabs,tabularx,siunitx}
\usepackage{svg}
\usepackage[pagenumbers]{cvpr} 

\usepackage[table]{xcolor}

\newcommand{\tit}[1]{\smallbreak\noindent\textbf{#1.}}
\newcommand{\tinytit}[1]{\noindent\textbf{#1.}}

\newcommand{\myverbatim}[1]{\texttt{\small\detokenize{#1}}}

\usepackage{nicefrac}
\usepackage{float}

\definecolor{cvprblue}{rgb}{0.21,0.49,0.74}
\usepackage[pagebackref,breaklinks,colorlinks,allcolors=cvprblue]{hyperref}
\usepackage[capitalize]{cleveref}

\def\paperID{18087} 
\def\confName{CVPR}
\def\confYear{2026}

\title{Robust and Calibrated Detection of Authentic Multimedia Content}

\author{
Sarim Hashmi$^{1}$ \quad
Abdelrahman Elsayed$^{1}$ \quad
Mohammed Talha Alam$^{1}$ \\
Samuele Poppi$^{1}$ \quad
Nils Lukas$^{1}$ \\
{\small $^{1}$Mohamed bin Zayed University of Artificial Intelligence (MBZUAI)}\\
{\tt\small \{sarim.hashmi, abdelrahman.elsayed, mohammed.alam, samuele.poppi, nils.lukas\}@mbzuai.ac.ae}
}

\begin{document}
\maketitle
\begin{abstract}
Generative models can synthesize highly realistic content, so-called \emph{deepfakes}, that are already being misused at scale to undermine digital media authenticity. 
Current deepfake detection methods are unreliable for two reasons: (i) distinguishing inauthentic content post-hoc is often impossible (e.g., with memorized samples), leading to an unbounded false positive rate (FPR); and (ii) detection lacks robustness, as adversaries can adapt to known detectors with near-perfect accuracy using minimal computational resources. 
To address these limitations, we propose a \emph{resynthesis} framework to determine if a sample is authentic or if its authenticity can be plausibly denied. 
We make two key contributions focusing on the high-precision, low-recall setting against efficient (i.e., compute-restricted) adversaries. 
First, we demonstrate that our calibrated resynthesis method is the most reliable approach for verifying authentic samples while maintaining controllable, low FPRs. 
Second, we show that our method achieves adversarial robustness against efficient adversaries, whereas prior methods are easily evaded under identical compute budgets. 
Our approach supports multiple modalities and leverages state-of-the-art inversion techniques.
\end{abstract}    
\section{Introduction}
\label{introduction}

Recent advances in large-scale generative models have led to notable improvements in the photorealism of synthesized media across modalities.
Text-to-image and diffusion families (e.g., DALL-E variants~\cite{ramesh2021zero} and latent diffusion models~\cite{rombach2022high}) now produce images that are frequently hard to distinguish from real images; similar progress has rapidly followed in audio~\cite{defossez2023audiocraft} and video generation~\cite{openai2024sora} 
These advances enable valuable creative and assistive applications, but they also erode a core assumption of digital forensics: namely, that post-hoc analysis can reliably separate authentic from synthetic content. 
The rise of highly capable, fine-tuned generators, and the ease with which adversaries can adapt them, therefore poses two linked challenges for content authenticity: (i) resynthesis indistinguishability (generators can reproduce real content with extremely high fidelity) and (ii) robustness (small, imperceptible manipulations can defeat many detectors).
As a result, simply asking \textit{“is this image fake?”} is no longer a reliable approach. 

Existing defenses fall into two categories: watermarking-based methods and post-hoc detection.
watermarking adds authenticity signals during generation but requires model changes and is vulnerable to removal and distribution-level attacks.
Post-hoc detectors instead look for residual artifacts left by generators; while effective against earlier GANs~\cite{goodfellow2014generative}, these methods suffer from brittle generalization to new image generation methods~\cite{fakedetect1,fakedetect2,fakedetect3,fakedetect4,fakedetect5,fakedetect6}, high false positive/negative rates on in-the-wild data~\cite{false_positive}, and poor robustness to adversarial perturbations and common image transformations~\cite{cocchi2023unveiling}. For many applications, especially those requiring high trust, it is important to bound the false positive rate (FPR); however, doing so with traditional detectors is often infeasible unless we operate in a high-precision, low-recall regime.
%

We therefore propose an alternative approach: moving from binary \emph{real vs.\ fake} detection to a calibrated notion of authentic vs.\ plausibly deniable.
%
An image is considered authentic only if its origin can be confidently verified as real; otherwise either because it is synthetic or because it can be faithfully resynthesized by existing generators it falls into the plausibly deniable region.
Many images cannot be definitively labeled as fake or authentic. Instead of forcing binary decisions, we aim to provide calibrated, interpretable risk scores.

We introduce an authenticity index computed from inversion and resynthesis statistics that quantify how  a modern generator can reproduce an input. 
An image's authenticity is labeled ''plausibly deniable" if generator inputs can be found that reproduce its content with high perceptual similarity, meaning its authenticity cannot be established post-hoc regardless of the true origin. 
Our index aggregates complementary perceptual metrics (pixel-level, structural, and semantic similarity) to produce a more robust calibrated separation between easily-inverted and hard-to-invert content.

\begin{figure*}[htp]
  \centering
  \includegraphics[width=0.99\textwidth]{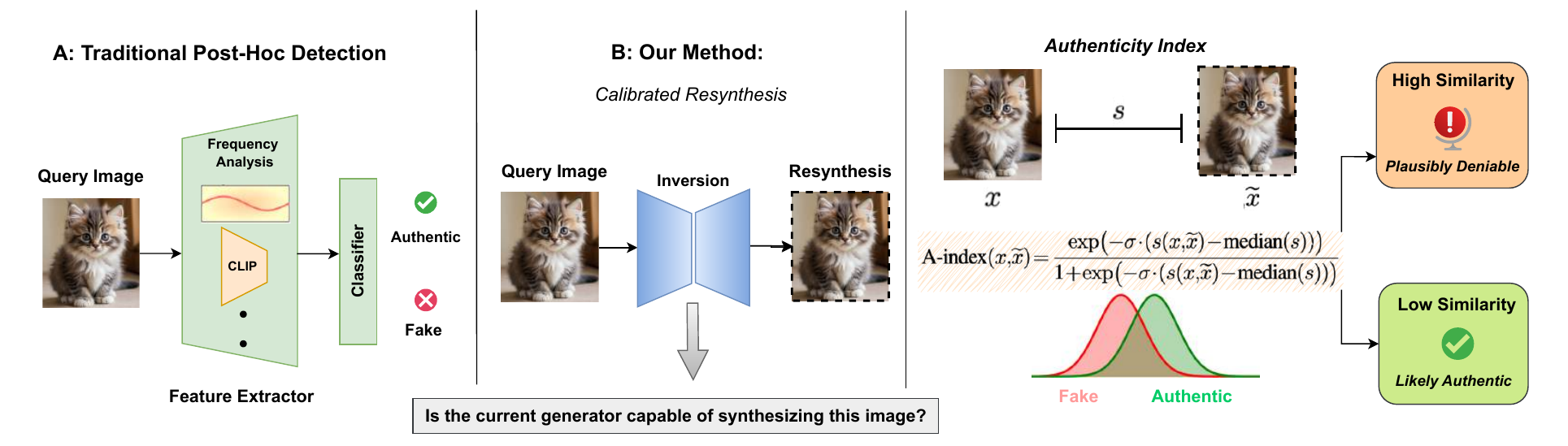}
  \caption{
\textbf{Conceptual illustration of the Authenticity Index.} 
(A) Traditional post-hoc detectors separate real from fake using feature cues but struggle as generators improve. 
(B) Our method instead asks: can a generator resynthesize the query image $x$? 
The similarity $s(x,\tilde{x})$ between $x$ and its inversion $\tilde{x}$ is calibrated into the \emph{Authenticity Index}, where high similarity implies \emph{plausible deniability} and low similarity indicates \emph{likely authentic}.
This allows us to reliably identify authentic content the generator is unlikely to have generated. 
}

  \label{fig:main-fig}
\end{figure*}

We adopt reconstruction-free inversion~\cite{rf_inversion} and we explicitly evaluate robustness under adversarial perturbations tailored to the inversion pipeline. 

Empirically, we show that this inversion-based authenticity index (i) generalizes better than several recent detectors under realistic distribution shifts, (ii) is substantially more robust to targeted adversarial perturbations that break conventional detectors, and (iii) is modality-agnostic, with an extension demonstrated on videos. 
%
%
As a key part of our evaluation, we validate the method on a large-scale social media corpus (~3,000  images scraped from public Reddit communities), demonstrating the real-world prevalence of plausibly deniable content and highlighting the practical risks of post-hoc authenticity verification.

In this work, we aim to bridge the gaps in understanding and improving the detection of synthetic content. Our key contributions are as follows:

\begin{itemize}
    \item We formalize \emph{plausible deniability} as an operational notion of authenticity, and introduce an authenticity index that operationalizes this paradigm through modern inversion and resynthesis.
    \item We propose adversarial objectives tailored for inversion pipelines and evaluate robustness under realistic threat model.
    \item We evaluate at scale, including a social-media study and an video modality extension, and show superior generalization and robustness compared to recent baselines.
\end{itemize}

Together, these contributions offer a principled, practically useful alternative to brittle binary detectors: instead of chasing an absolute “fake/real” label that may not exist, we provide a calibrated assessment of whether an image’s authenticity can be established post-hoc.

\section{Related Work}
Detection of AI-generated images has been approached through \textbf{passive forensic detection} and \textbf{latent inversion techniques}. In this section, we review key prior work in each area, highlighting their strengths, limitations, and robustness issues.
\subsection{Post-hoc Detection Methods}
Post-hoc detection methods analyze subtle forensic artifacts without altering image content. Early works utilized spectral anomalies and sensor noise fingerprints, such as Photo-Response Non-Uniformity (PRNU)~\cite{frank2020leveraging,lukas2006digital}. Recent approaches primarily employ deep learning models trained on large datasets, notably FaceForensics++~\cite{rossler2019faceforensics++} and the DeepFake Detection Challenge (DFDC)~\cite{dolhansky2020deepfake}. Models based on architectures like XceptionNet and vision transformers have significantly advanced detection capabilities~\cite{rossler2019faceforensics++}. However, these methods often suffer from high false-positive rates and poor generalization to unseen generative architectures~\cite{cozzolino2024zero}. Furthermore, passive detectors are vulnerable to adversarial attacks and image transformations commonly employed on social media, challenging their real-world applicability~\cite{frank2020leveraging}.

Emerging zero-shot methods, such as ZED aim to generalize better by training only on real images, thus providing a promising direction for handling unknown generative models~\cite{cozzolino2024zero}. Despite these improvements, the evolving nature of generative models and adversarial strategies continuously challenges passive detection efficacy.

\subsection{Inversion-Based Detection Methods}
Inversion-based detection approaches reconstruct suspect images into a generative model’s latent space to assess authenticity. This concept leverages the notion that synthetic images typically reconstruct with lower errors compared to real images~\cite{zhu2020domain}. Methods like LatentTracer formalize inversion-based detection by leveraging gradient-based optimization to distinguish images generated by specific diffusion models from real images or those generated by other models~\cite{wang2024trace}. Similarly, GAN inversion techniques have been employed to detect GAN-generated images based on latent-space reconstruction fidelity~\cite{zhu2020domain,zhu2018hidden}.

Despite promising results, inversion-based methods require direct model access, limiting their scalability across diverse or unknown generative models. Additionally, overlapping distributions between real and synthetic images, as well as varying scene complexity, can introduce significant false positives or negatives, complicating their practical implementation~\cite{wang2024trace,zhu2020domain}.

Collectively, these diverse methodologies illustrate substantial progress but also underscore persistent challenges in reliably identifying AI-generated imagery. Our work aims to bridge these gaps, proposing an integrated framework that addresses the inherent limitations of existing methods to enhance detection robustness in real-world scenarios.

\section{Methodology}
\label{sec:methodology}


\subsection{Preliminaries}

In this section, we review the core technical ingredients that underpin our Authenticity Index (A-index), by introducing foundational concepts like \textit{diffusion generative modeling}, \textit{latent‑space inversion }, and the \textit{reconstruction‑free inversion paradigm}. 

\tit{Diffusion Generative Modeling} Diffusion models~\cite{ho2020denoising} construct a forward stochastic noising process and a learned reverse denoising process that together transform simple Gaussian noise into complex images. Concretely, given a data sample \(x_0 \sim p_{\mathrm{data}}\), the forward process defines
\[
x_t = \sqrt{1-\beta_t}\,x_{t-1} + \sqrt{\beta_t}\,\epsilon_{t-1}, 
\quad \epsilon_{t-1} \sim \mathcal{N}(0,I),
\]
for \(t=1,\dots,T\), where \(\{\beta_t\}\) is a prescribed variance schedule~\cite{ho2020denoising}. The generative reverse process is parameterized by a neural network \(\epsilon_\theta(x_t,t)\) and models:
\[
p_\theta(x_{t-1}\mid x_t) 
= \mathcal{N}\bigl(x_{t-1}; \mu_\theta(x_t,t),\,\Sigma_\theta(x_t,t)\bigr),
\]
with \(\mu_\theta\) and \(\Sigma_\theta\) chosen so that sampling from \(p_\theta\) iteratively from \(x_T\sim\mathcal{N}(0,I)\) recovers \(x_0\) in distribution. These models have achieved photorealism on par with GANs while offering exact likelihoods and stable training dynamics~\cite{rombach2022high}. Importantly for our work, the latent trajectory $\{x_t\}$ admits \emph{inversion}, enabling techniques that map a given image back to an initial noise $z$--a property we will leverage in the next subsection on reconstruction-free inversion.

\tit{Reconstruction-Free Inversion for Authenticity Assessment} Among inversion methods, we focus on \emph{reconstruction-free inversion}~\cite{rf_inversion} (RF-Inversion), which quantifies a generator’s ability to reproduce an image’s defining features without performing full pixel-level reconstruction. Let $\Psi$ denote a feature extractor (\eg, Fourier magnitude spectra or wavelet coefficients). Given an image $x$, RF-Inversion applies an approximate inverter $\widetilde{G}^{-1}$ (which may be a shallow encoder). Formally, we define the feature discrepancy as
\[
\Delta\Psi = \bigl\|\Psi(x) - \Psi\bigl(\widetilde{G}^{-1}(x)\bigr)\bigr\|.
\]
A small $\Delta\Psi$ indicates that the generator can replicate the image’s core features, suggesting reduced authenticity. By focusing on feature distances rather than pixel reconstruction error, RF-Inversion avoids costly gradient-based optimization in pixel space and dramatically reduces computation~\cite{wang2024taming}. This makes it well suited for large-scale screening of images to assess whether they are likely produced by a given generative model. We employ different techniques to invert images based on the level of interpretability and knowledge of the models.


\tit{Similarity Metrics for Authenticity Assessment} To quantify the similarity between an image and its inverted counterpart, we rely on standard perceptual and signal-based metrics, like PSNR~\cite{psnr}, SSIM~\cite{ssim}, LPIPS~\cite{lpips}, and CLIP similarity~\cite{clip} , The corresponding formulas are provided in the supplementary material.

\begin{figure}[ht]
    \centering
    \includegraphics[width=0.8\linewidth]{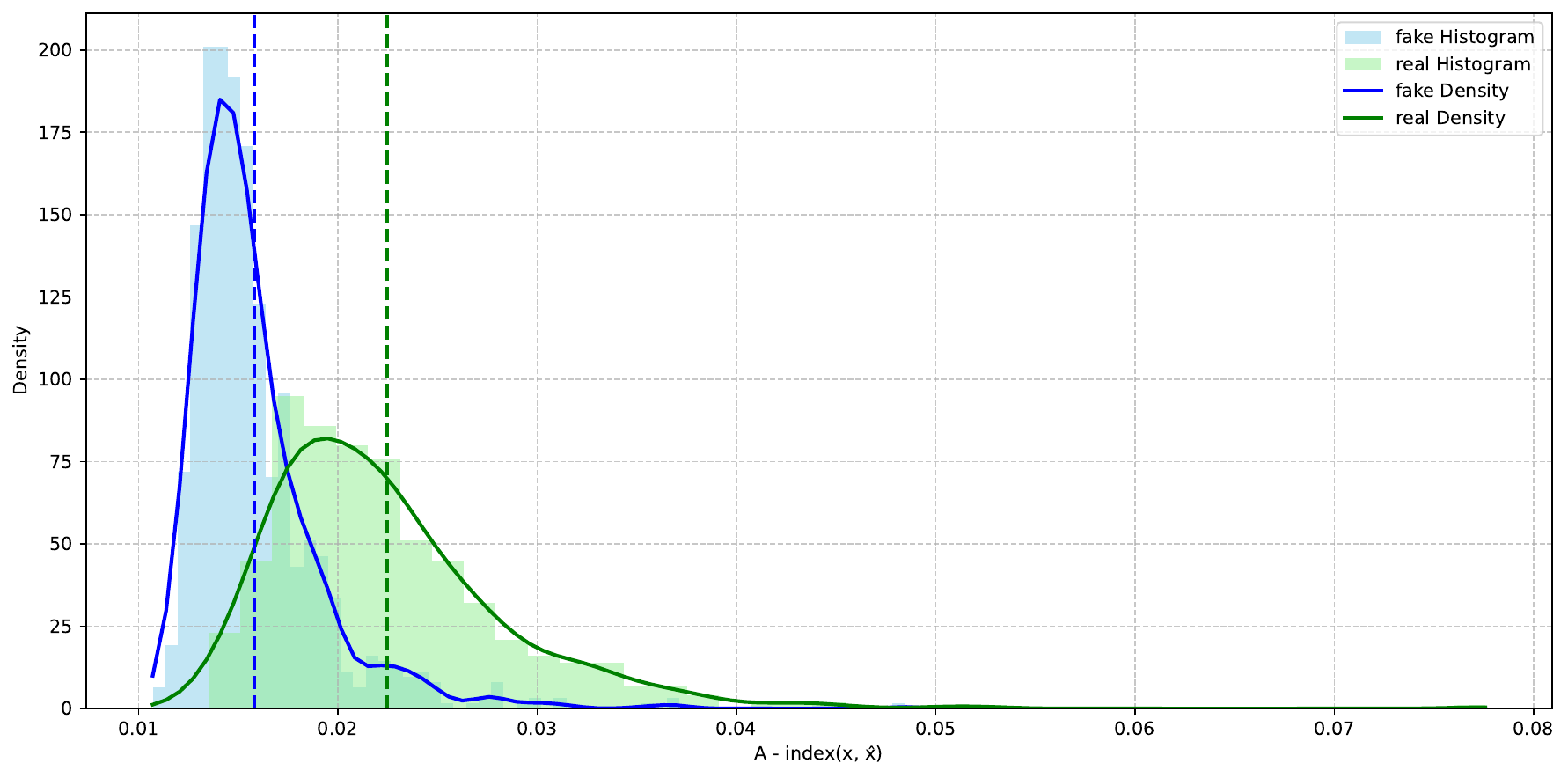}
    \caption{Distribution
    of A-index$(x,\tilde{x})$ for fake image and real image.}
    \label{fig:methodology_a_calib}
\end{figure}

\subsection{From Binary Detection to Calibrated Authenticity}

Traditional deepfake detection approaches treat authenticity as a binary decision: an image is either real or fake (~\Cref{fig:main-fig}). However, as discussed in~\Cref{introduction}, this framing is increasingly unreliable: some real images invert poorly and appear synthetic, while some generated images invert easily and appear real. 

We propose a calibrated notion of authenticity that reflects this ambiguity. Instead of forcing every image into a binary label, we distinguish between (1) \textbf{Authentic:} images that remain consistently stable under inversion, producing reconstructions above a calibrated threshold, (2) \textbf{Plausibly deniable:} all other cases, including generated images and real images whose inversion yields reconstructions that are too easily reproduced by the generator.

\begin{figure*}[t]
    \centering
    \includegraphics[width=0.9\linewidth]{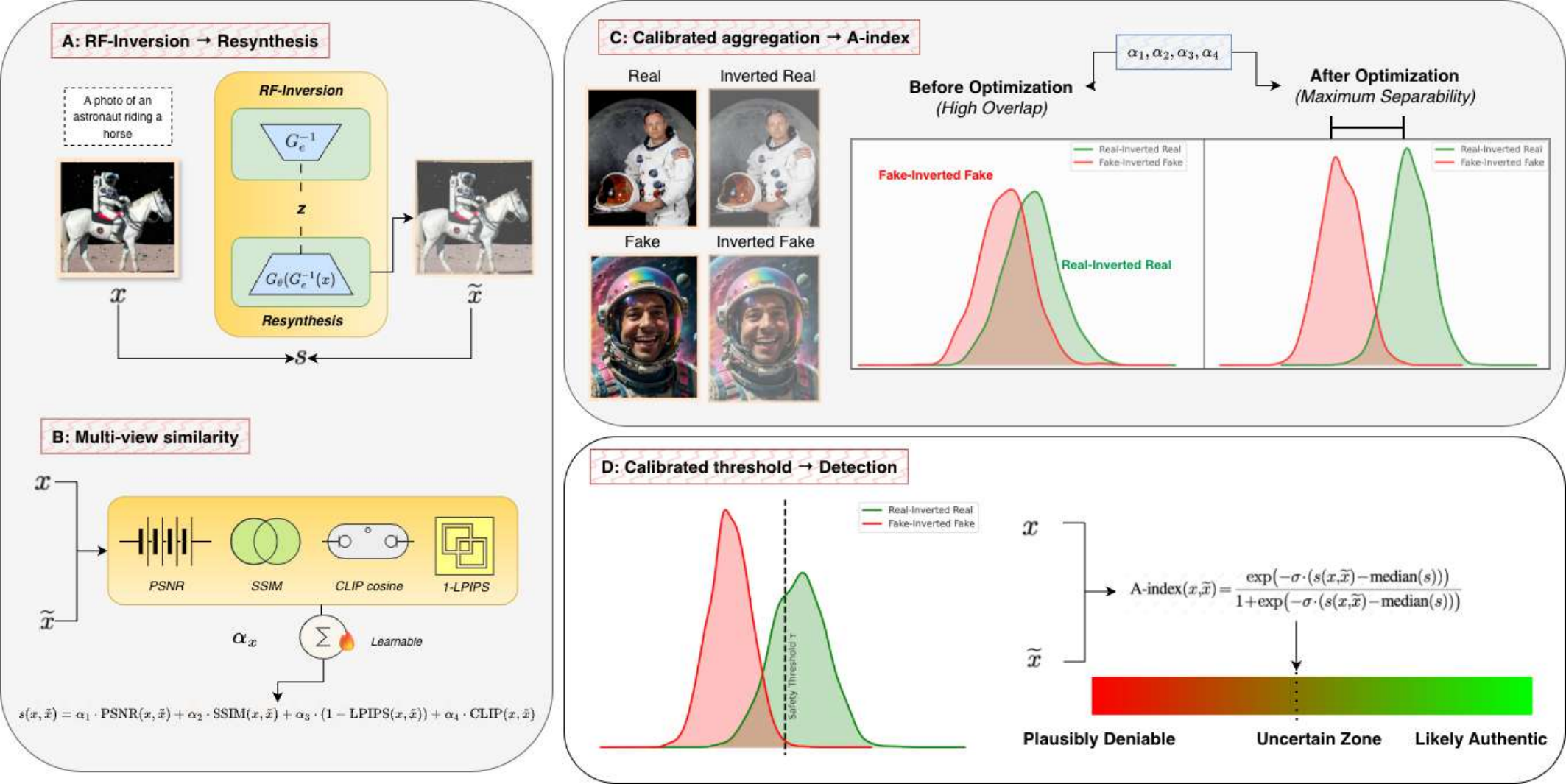}
    \caption{\textbf{Architecture of the Authenticity Index (A-Index).} Given an input image $x$, a reconstruction-free inverter $G_e^{-1}$ produces an inverted reconstruction $\tilde{x}$. We then compute complementary similarities between $x$ and $\tilde{x}$: pixel fidelity\texttt{(PSNR)}, structural fidelity\texttt{(SSIM)}, perceptual distance\texttt{($1{-}$LPIPS)}, and semantic consistency (\texttt{CLIP} cosine). A calibrated weighted combiner (learned $\alpha_1, \alpha_2, \alpha_3, \alpha_4$) produces a scalar $s$($x$, $\tilde{x}$), yielding the A-Index in $[0,1]$. A safety threshold $\tau$ certifies content as \emph{Authentic} when $\text{A-Index} \geq \tau$, and otherwise labels it as \emph{Plausibly Deniable}. We further analyze robustness by applying $\ell_\infty$-bounded perturbations $\delta$ (PGD-style) through $G_e^{-1}$ to maximize or minimize the A-Index.
}
\label{fig:concept}
\end{figure*}

This framing aligns with the operational limits of post-hoc forensics: if a generator can reproduce an input with high perceptual fidelity, the content’s authenticity cannot be guaranteed, regardless of its true origin. Our methodology therefore aims not to declare an absolute binary verdict, but to provide a calibrated score that separates authentic from plausibly deniable content.

\subsection{Authenticity Index}

Having introduced RF-Inversion and similarity metrics, we now define the \emph{Authenticity index}, a composite score designed to quantify how well an image aligns with its inverted counterpart. The index is built as a weighted combination of complementary image quality measures, ensuring that both low-level fidelity and high-level semantic consistency are captured.

Formally, given an image $x$ and its inverted reconstruction $\widetilde{x}$, we compute four similarity measures: PSNR, SSIM, inverted LPIPS (\ie, $1-\text{LPIPS}$), and CLIP similarity. These are aggregated into a linear combination
\begin{equation}
\begin{split}
s(x,\tilde{x}) =\;& \alpha_1 \cdot \text{PSNR}(x,\tilde{x}) + \alpha_2 \cdot \text{SSIM}(x,\tilde{x}) \\
&+ \alpha_3 \cdot (1 - \text{LPIPS}(x,\tilde{x})) + \alpha_4 \cdot \text{CLIP}(x,\tilde{x}).
\end{split}
\end{equation}

where $\alpha_1,\dots,\alpha_4$ are scalar weights. To normalize and bound the score, we apply a sigmoid transformation with scale parameter $\sigma$:
\[
\text{A-index}(x,\widetilde{x}) = \frac{\exp\bigl(-\sigma \cdot (s(x,\widetilde{x})  \big)}{1 + \exp\bigl(-\sigma \cdot (s(x,\widetilde{x}) \bigr)}.
\]

The weights $\alpha_1,\dots,\alpha_4$ are selected through an optimization procedure that minimizes the overlap between score distributions of real vs.\ inverted real images and fake vs.\ inverted fake images. To do that, we adopt differential evolution, a global optimization algorithm that allows both positive and negative weights to capture different relationships between similarity metrics and inversion quality(  \Cref{fig:methodology_a_calib}).

This authenticity index operationalizes the plausibly deniable paradigm: it provides a calibrated measure of how well an image aligns with its inversion, serving as the primary metric in our experiments to distinguish content that can be asserted as authentic from content that remains plausibly deniable.
\Cref{fig:concept} shows the entire pipeline of our $\text{A-index}(x,\widetilde{x})$ calibration.

\subsection{From Authenticity Index to Detection}

While the authenticity index provides a calibrated score of inversion consistency, using it for detection requires reasoning over the distributions of scores assigned to real and generated images. Concretely, given a dataset with both classes, we compute the authenticity index for each image, yielding two empirical distributions: one for real images and one for generated images.

We observe three key properties of these distributions: (1) Variability, where both real and generated images span a range of scores: some generated images achieve moderately high values, while real images vary depending on how well they align with the inversion process, (2) Importantly, real images that are difficult to invert tend to achieve the \emph{highest} scores, extending the tail of the real distribution beyond that of the fake distribution.(3) This separation enables the definition of a threshold~$\tau_{\text{safety}}$ such that any image with authenticity index $\text{A-index}(x,\widetilde{x}) \geq \tau_{\text{safety}}$ can be confidently classified as \emph{real} with 1\% False Positive Rate (FPR). Images below $\tau_{\text{safety}}$ are not necessarily fake, but fall into the \emph{plausibly deniable} region.

In this way, the authenticity index underpins a practical detection method: instead of forcing every input into a binary real/fake label, we identify a calibrated boundary that guarantees the authenticity of content above threshold while acknowledging uncertainty below it. This procedure makes explicit the transition from a metric of inversion fidelity to an operational detection rule.


\subsection{Robustness to Adversarial Attacks}

Beyond serving as a static metric, the authenticity index also enables robustness analysis: we can probe its stability under adversarial perturbations of the input image. This motivates the following study of adversarial robustness.

\tit{Motivation} Conventional adversarial attacks such as FGSM~\cite{goodfellow2014explaining} and PGD~\cite{madry2017towards} are designed for classification pipelines and rely on perturbing logits with simple loss functions. However, our setting involves a generative inversion pipeline, where direct application of these methods is meaningless. 

\tit{Threat Model} We therefore define a task-specific threat model: given an image $I$, the adversary seeks to add a small perturbation $\delta$ (bounded by $\ell_\infty$ norm) such that the inversion process $\widetilde{G}^{-1}(I+\delta)$ alters the Authenticity Index. Importantly, this shift in authenticity enables a new paradigm of analysis: instead of the binary distinction \emph{real vs.\ fake}, we identify the regime of \emph{real vs.\ uncertain}. Real images that remain stable above a threshold are deemed authentic, while those driven below it become indistinguishable from generated content.

\tit{Adversarial Objective for Inversion}
\label{sec:adversial_attack} Formally, we define the adversarial optimization problem as
\[
\max_{\|\delta\|_{\infty} \leq \epsilon} \ \text{A-index}\bigl(I, \widetilde{G}^{-1}(I+\delta)\bigr),
\]
where $\mathcal{AI}$ denotes our authenticity index that aggregates PSNR, SSIM, inverted LPIPS, and CLIP similarity into a calibrated score, and $\epsilon = 8/255$ constrains the perturbation magnitude to remain imperceptible to the human eye. Depending on the optimization direction, one can also minimize the objective, thereby reducing the authenticity score of real images.

The adversarial perturbation $\delta$ is optimized via gradient-based methods over a differentiable approximation of the inversion pipeline, while the $\ell_\infty$ constraint is enforced after each step. Unlike classification-based attacks, the gradient flows through the entire generative inversion process rather than through classification logits.

\vspace{0.5em}

This formulation provides a principled way to evaluate the robustness of the authenticity index under adversarial perturbations. In the experimental section, we instantiate this framework to quantify robustness margins across real and generated datasets.
To account for adversarial vulnerability, we additionally define a new threshold, $\tau_{\text{security}}$, such that any image with authenticity index $\text{A-index}(x,\widetilde{x}) \geq \tau_{\text{security}}$ can be confidently classified as \emph{real} with a 1\% False Positive Rate (FPR) in an adversarial attack setting.




\begin{table}[h]
\caption{Results before/after PGD ($\epsilon=\frac{8}{255}$) on 2{,}000 images showing correctly classified fake/real images, accuracies, and attack success rates.}
\label{tab:attack_results}
\setlength{\tabcolsep}{2pt}
    \centering
        \resizebox{\linewidth}{!}{%
        \begin{tabular}{l ccc ccc cc}
        \toprule
        & \multicolumn{3}{c}{\textbf{Before attack}}
        & \multicolumn{3}{c}{\textbf{After attack}}
        & \multicolumn{2}{c}{\textbf{Attack success (\%)}} \\
        \cmidrule(lr){2-4}\cmidrule(lr){5-7}\cmidrule(lr){8-9}
        \textbf{Model}
        & \shortstack{Correct\\fake}
        & \shortstack{Correct\\real}
        & \shortstack{Acc.\\(\%)}
        & \shortstack{Correct\\fake}
        & \shortstack{Correct\\real}
        & \shortstack{Acc.\\(\%)}
        & \shortstack{on\\fake}
        & \shortstack{on\\real} \\
        \midrule
        UFD~\cite{ufd}       & 1   & 974 & 48.75 & 0  & 0  & \cellcolor{red!12}\textbf{0.00} & 100.0 & 100.0 \\
        FreqNet\cite{fakedetect6}  & 53  & 995 & 52.40 & 0  & 0  & \cellcolor{red!12}\textbf{0.00} & 100.0 & 100.0 \\
        NPR~\cite{npr}       & 86  & 653 & 36.95 & 0  & 0  & \cellcolor{red!12}\textbf{0.00} & 100.0 & 100.0 \\
        FatFormer~\cite{fatformer} & 1   & 987 & 49.40 & 0  & 0  & \cellcolor{red!12}\textbf{0.00} & 100.0 & 100.0 \\
        D3~\cite{d3}         & 736 & 942 & \cellcolor{green!15}83.90 & 24 & 11 & \cellcolor{yellow!20}1.75 & 96.7  & 98.8  \\
        C2PClip~\cite{c2pclip}             & 51  & 949 & 50.00 & 0  & 2  & \cellcolor{red!12}\textbf{0.10} & 100.0 & 99.8  \\
        \bottomrule
        \end{tabular}
        }
        \vspace{-.35cm}
\end{table}

\section{Experiments}

\subsection{Setting}
We evaluate our method using five open-weight generative models: Stable Diffusion 2.1~\cite{rombach2021ldm}, Stable Diffusion 3 (medium)~\cite{stability_sd3_medium,esser2024sd3}, Stable Diffusion 3.5 (medium)~\cite{stability_sd35}, FLUX.1 [dev]~\cite{flux2024}, and FLUX.1 [dev] with a Realism LoRA adapter~\cite{xlabs_flux_realismlora}. For evaluation, we employ a curated subset of the De-Factify 4 dataset~\cite{defactify4}, which contains 1{,}000 real images and 1{,}000 corresponding synthetic images, with 200 images \emph{generated} from each of five generation methods present in the dataset (Stable Diffusion 2.1, Stable Diffusion XL, Stable Diffusion 3, DALL·E 3, and Midjourney 6)~\cite{malviya2025skdu}. All experiments are conducted using Rectified-Flow Inversion (RF-Inversion)~\cite{rout2025rfinversion} executed on 4$\times$ NVIDIA A6000 GPUs (50\,GB each), which enables large-scale inversion across models and samples. The inversion process hyperparameters are provided in the supplementary material. 
We also evaluate six baselines alongside our method in a zero-shot setting. The six baselines are UFD~\cite{fakedetect3}, FreqNet~\cite{fakedetect6}, NPR~\cite{npr}, FatFormer~\cite{fatformer}, D3~\cite{d3}, and C2PClip~\cite{c2pclip}.For the social-media study, we further collect $\sim$3{,}000 images from Reddit using diverse queries to evaluate the prevalence and risk patterns of plausibly deniable content in the wild. More details are provided in the supplementary material.

\subsection{Results on Generalizability and Robustness to Adversarial Attacks}
\begin{figure}[t]
    \centering
    \includegraphics[width=\linewidth]{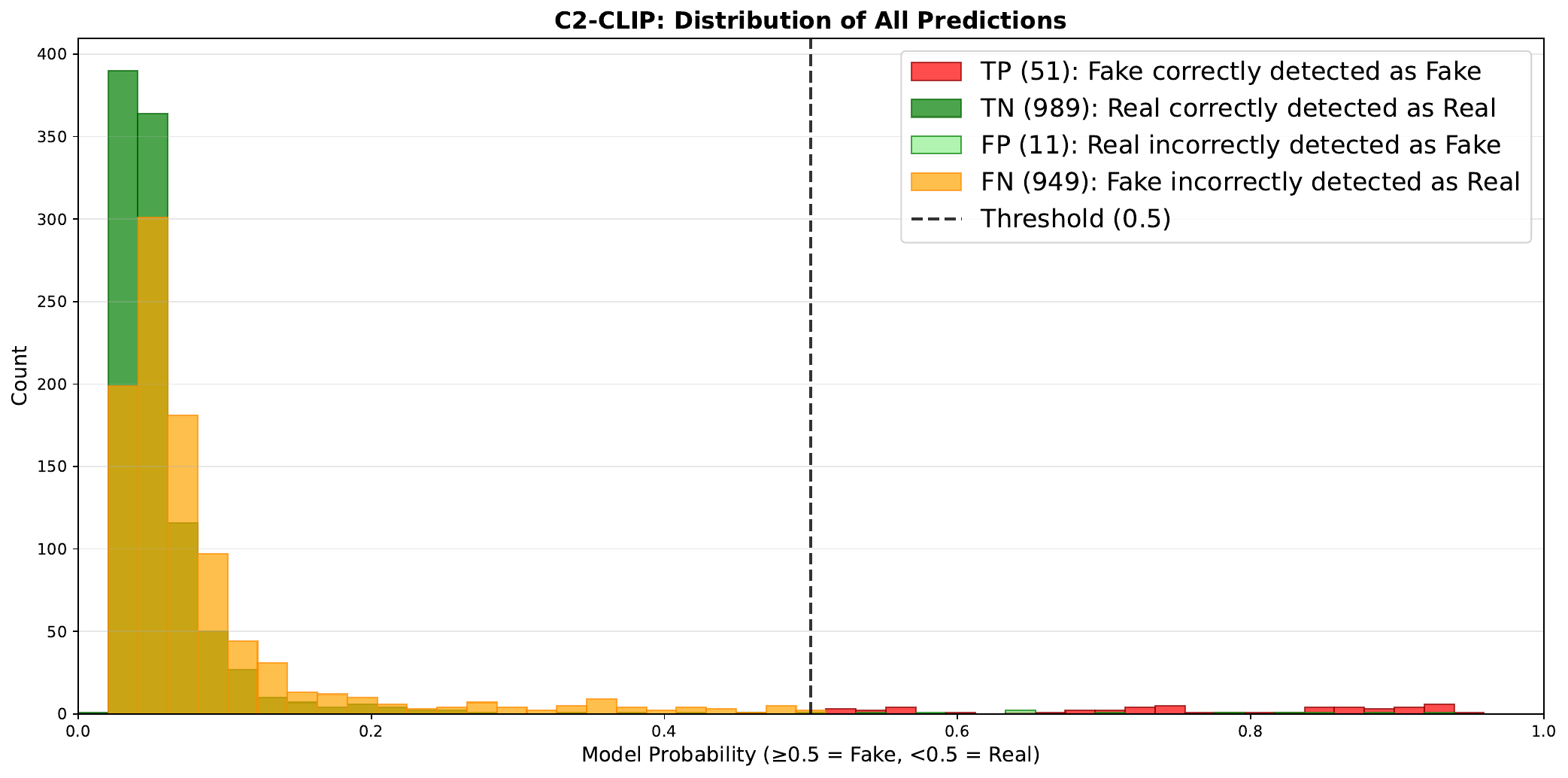}
\caption{Like most traditional methods, C2-CLIP fails to generalize, as real and fake prediction densities heavily overlap, indicating poor separability.}

    \label{fig:c2_clip}
\end{figure}

\begin{figure}[b]
    \centering
    \includegraphics[width=\linewidth]{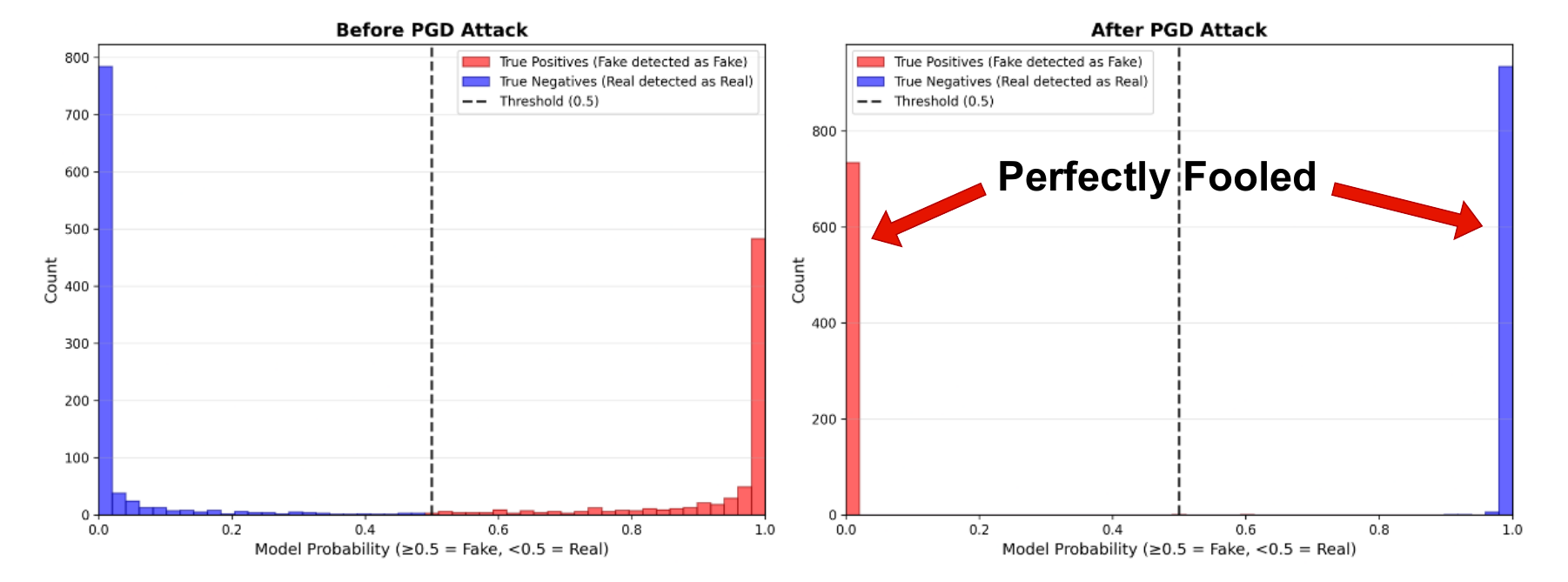}
\caption{Robustness evaluation of the traditional D3 detector under PGD attack. Before the attack, the model confidently distinguishes between real and fake samples. However, after the adversarial perturbation, the distribution inverts completely, resulting in near-total misclassification , the detector is effectively and perfectly fooled.}
    \label{fig:d3_pgd}
    
\end{figure}

\tit{Current Deepfake Detectors do not Generalize} Most existing deepfake detectors rely on binary classification to distinguish between real and fake images. However, as shown in ~\Cref{fig:c2_clip}, this approach often leads to a strong bias toward classifying out-of-domain inputs as real. In particular, the C2-CLIP detector correctly classifies only 51 fake images, while misclassifying 949 as real, despite correctly identifying most authentic inputs. This imbalance highlights a strong bias toward high recall on real images at the cost of failing to flag synthetic content. A similar failure pattern is evident under adversarial conditions. As shown in~\Cref{fig:d3_pgd} and~\Cref{tab:attack_results}, all evaluated detectors collapse completely under imperceptible PGD~\cite{pgd} perturbations, with several models including UFD~\cite{fakedetect3}, FreqNet~\cite{fakedetect6}, and FatFormer~\cite{fatformer} dropping to 0\% accuracy post-attack. Even D3~\cite{d3}, the strongest baseline, falls from 83.90\% to just 1.75\% accuracy. These results underscore a general and critical weakness: traditional detectors exhibit a strong confirmation bias toward authenticity and offer no graceful degradation under adversarial manipulation or distributional shift. Refer to supplementary for additional results.

To address these limitations, our framework adopts an opposing philosophy: instead of forcing a binary decision, we prioritize \textbf{high precision and low recall} for authentication. We confidently certify only those images whose authenticity can be established with certainty, while abstaining from judgment on uncertain cases (\emph{Plausible Deniablity}). This design reflects the practical reality that a false authenticity claim is far more damaging than a conservative abstention.

\tit{Adversarial Robustness} We evaluate whether adversaries can push fake images above the threshold~$\tau$ using imperceptible perturbations. Unlike binary classifiers that require only crossing a decision boundary, our threshold-based framework defines two calibrated cutoffs: a \emph{Safety Threshold}~$\tau_{\text{safety}}$ that guarantees low false positives on authentic data, and a \emph{Security Threshold}~$\tau_{\text{security}}$ that limits false acceptances of adversarially perturbed fakes. Successful attacks must therefore not only invert the input convincingly, but also elevate the authenticity index past strict, data-driven thresholds making evasion substantially more difficult in practice ,Through this evaluation, we determined that for Stable Diffusion 3 (medium) the threshold $\tau_{\text{safety}}$ is 0.0365, while the corresponding adversarially calibrated threshold $\tau_{\text{security}}$ is 0.038.

\tinytit{Protocol} To assess generalization to unseen generators, we evaluate six baselines alongside our method in a zero-shot setting where test images are generated by models not seen during training or calibration. On top of that, we employ PGD attacks~\citep{pgd} with $\ell_\infty$ budget $\epsilon = \nicefrac{8}{255}$ (maximum 8 intensity levels per pixel) to assess the models in adversarial attacks setting. We report before-attack accuracy, post-attack accuracy, attack success rate, precision and recall.

\tinytit{Results} Existing detectors exhibit complete and catastrophic collapse under adversarial perturbations. \Cref{tab:attack_results} highlights this brittleness under PGD attacks with $\epsilon=\nicefrac{8}{255}$. UFD~\cite{ufd} degrades from 48.75\% accuracy to 0.00\%, misclassifying \emph{every} image after attack and achieving a 100.0\% attack success rate (ASR) on both fake and real samples. FreqNet~\cite{fakedetect6} suffers the same outcome, dropping from 52.40\% to 0.00\% accuracy with 100.0\% ASR. The trend continues across all baselines: NPR~\cite{npr}, FatFormer~\cite{fatformer}, and C2PClip~\cite{c2pclip} are fully compromised post-attack, with 100.0\% ASR on fakes and either 100.0\% (NPR, FatFormer) or 99.8\% (C2PClip) ASR on reals. Even D3~\cite{d3}, the strongest baseline, falls sharply from 83.90\% to just 1.75\% accuracy under attack.

\begin{figure}[t]
    \centering
        \includegraphics[width=\linewidth]{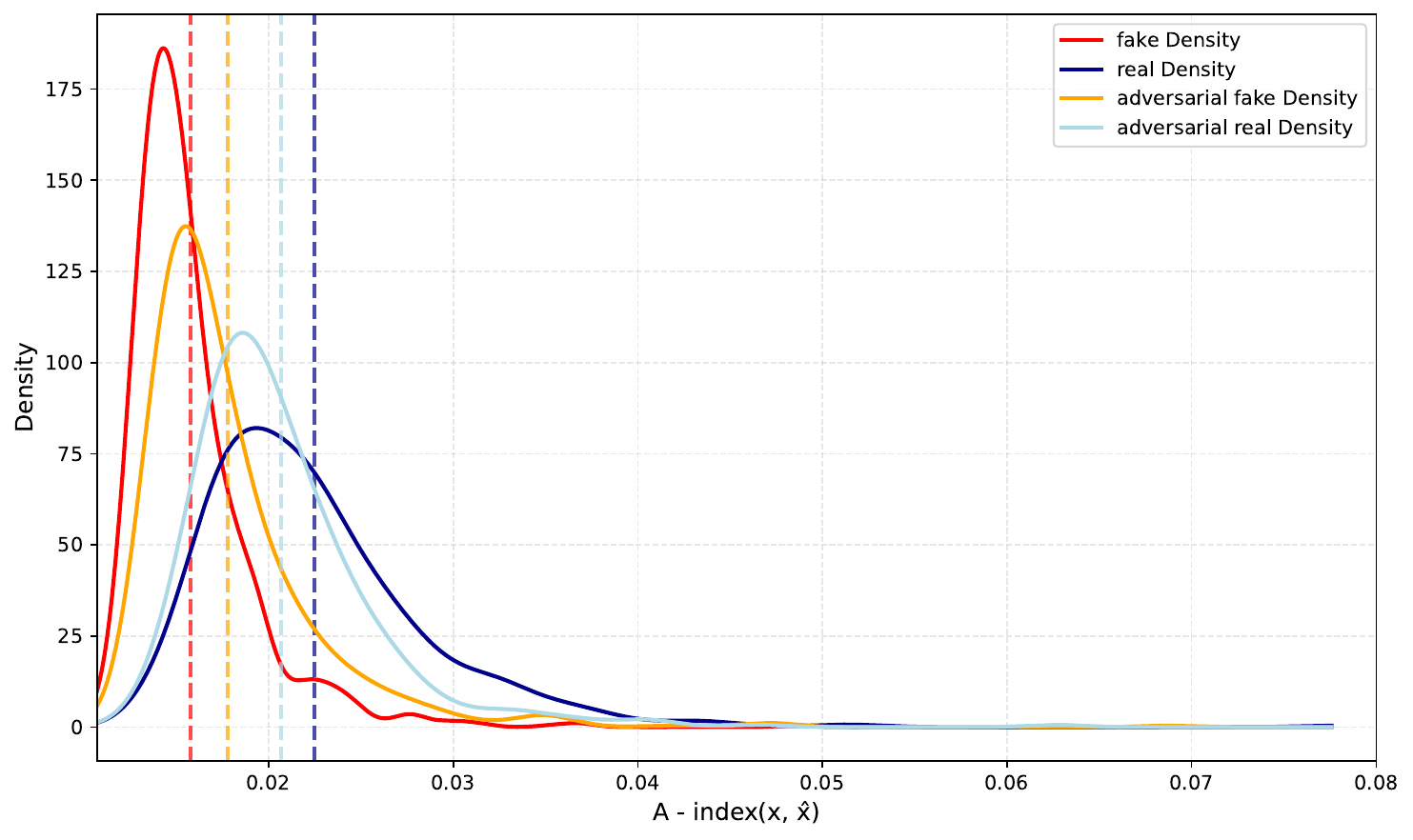}
        \caption{A-index density comparison across four settings for distribution comparison under adversarial perturbations.}
    \label{fig:distribution_comparison}

    \vspace{-.5cm}
\end{figure}

This brittleness stems from architectural limitations: binary classifiers must produce confident predictions on all inputs. When adversarial perturbations shift samples across the decision boundary, they induce complete failure with no mechanism for graceful degradation. In contrast, our method maintains significant robustness. As shown in~\Cref{fig:distribution_comparison}, even under perturbation, the A-index distributions for real and fake images remain well-separated. Adversarial examples increase overlap modestly but do not collapse the distribution. Our abstention-based framework therefore preserves utility where binary classifiers fail completely, offering a calibrated degradation path grounded in perceptual similarity rather than forced decisions.

To address these limitations, our framework adopts an opposing philosophy: instead of forcing a binary decision, we prioritize \textbf{high precision and low recall} for authentication. We confidently certify only those images whose authenticity can be established with certainty, while abstaining from judgment on uncertain cases (\emph{Plausible Deniablity}). This design reflects the practical reality that a false authenticity claim is far more damaging than a conservative abstention.

\subsection{Medium Resource Attacker}
\label{sec:medium-attacker}

We first note the limiting case: if an attacker had effectively infinite resources and could synthesize and invert millions of candidates, then they could select the candidate with maximal authenticity score and adversarially refine it to drive the defender’s metric arbitrarily high, in which case post-hoc verification would be defeated.

In practice, attackers are resource-limited; we therefore analyze a medium-resource protocol in which the attacker is given a single prompt, samples \(N\) random seeds, and synthesizes candidates \(x_i = G_\theta(z_i;\text{prompt})\) for \(i=1,\dots,N\). For each candidate the attacker obtains the inversion \(\widetilde{x}_i = \widetilde{G}^{-1}(x_i)\) and evaluates the authenticity index \(\textbf{A-index}(x_i,\widetilde{x}_i)\). The attacker then selects \(x^\star = x_{i^\star}\) with \(i^\star=\arg\max_{i=1,\dots,N}\mathcal{AI}(x_i,\widetilde{x}_i)\) and searches for a small perturbation \(\delta\) that increases the score when the perturbed image is passed into the inversion routine. Formally, the adversarial problem is written as

\[
\max_{\|\delta\|_\infty \le \varepsilon}\; \textbf{A-index}\bigl(x^\star,\; \widetilde{G}^{-1}(x^\star + \delta)\bigr),
\qquad \varepsilon = 8/255.
\]

\begin{figure}[b]
    \centering
    \includegraphics[width=\linewidth]{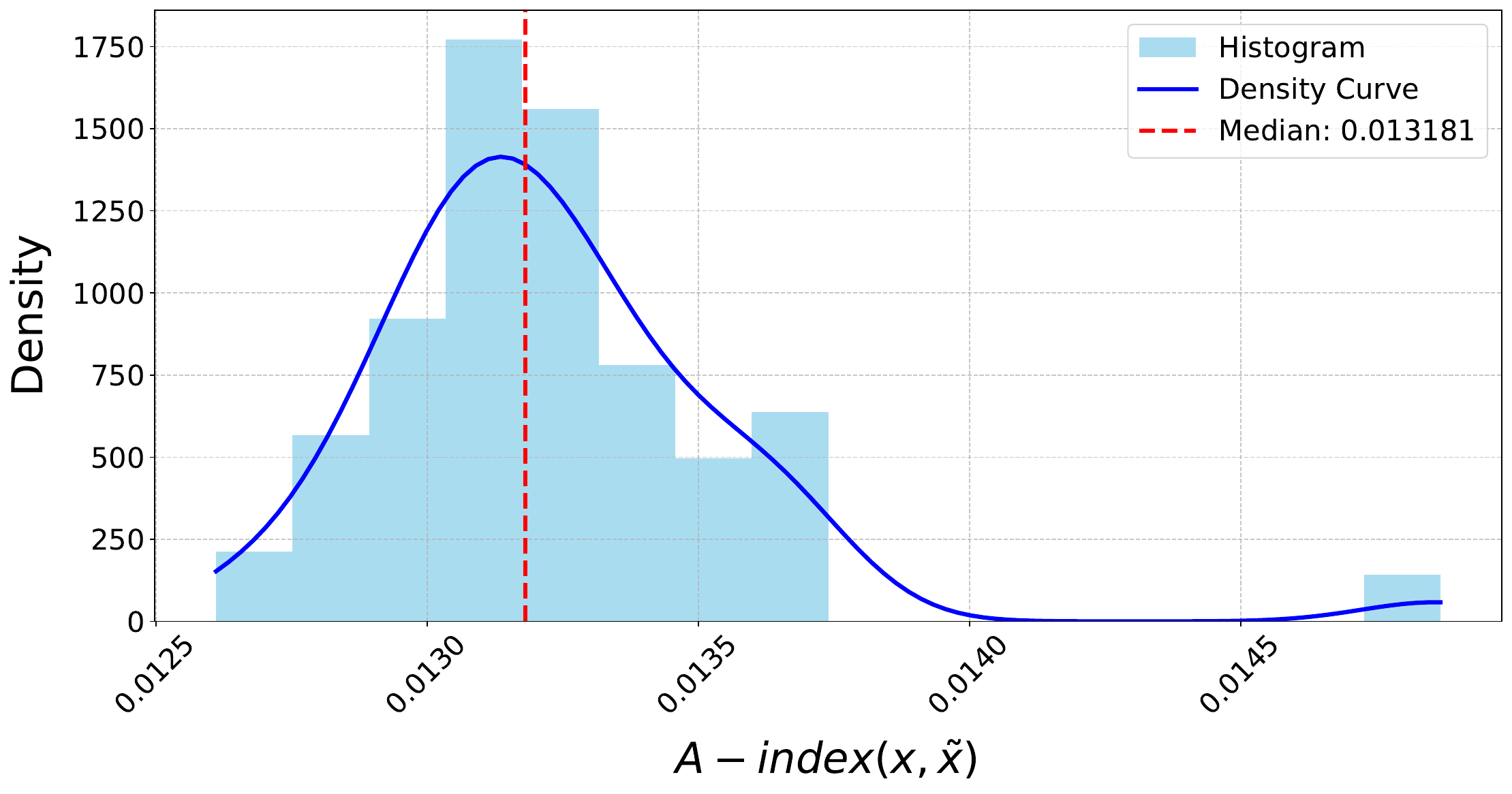}
    \caption{Distribution of $A-index(x,\tilde{x})$ for N=100.}
    \label{fig:medium_attacker}
\end{figure}

This objective is implemented with projected gradient ascent on \(\delta\): at each iteration the attacker computes or estimates gradients of a differentiable approximation of \(\widetilde{G}^{-1}\) and of \(\textbf{A-index}\) with respect to the inverter input, takes an ascent step on \(\delta\), projects \(\delta\) back to the admissible set \(\{\delta:\|\delta\|_\infty\le\varepsilon\}\), and clips \(x^\star+\delta\) to the valid pixel range. To characterize this attack we ran experiments with \(N=100\); the density of candidates versus \(\textbf{A-index}\) is plotted in~\Cref{fig:medium_attacker}. From the sampled pool we selected the top-scoring candidate whose \(\textbf{A-index}\) was 0.0148. After performing the constrained adversarial refinement described above the reported score increased only to 0.0154, indicating a small shift and suggesting that for this prompt and this budget the medium-resource attacker achieves only modest improvement and it is also below the safety threshold and security threshold.

\subsection{Social Media Study and the Effect of Adapters}

\begin{figure}[ht]
    \centering
    \includegraphics[width=\linewidth]{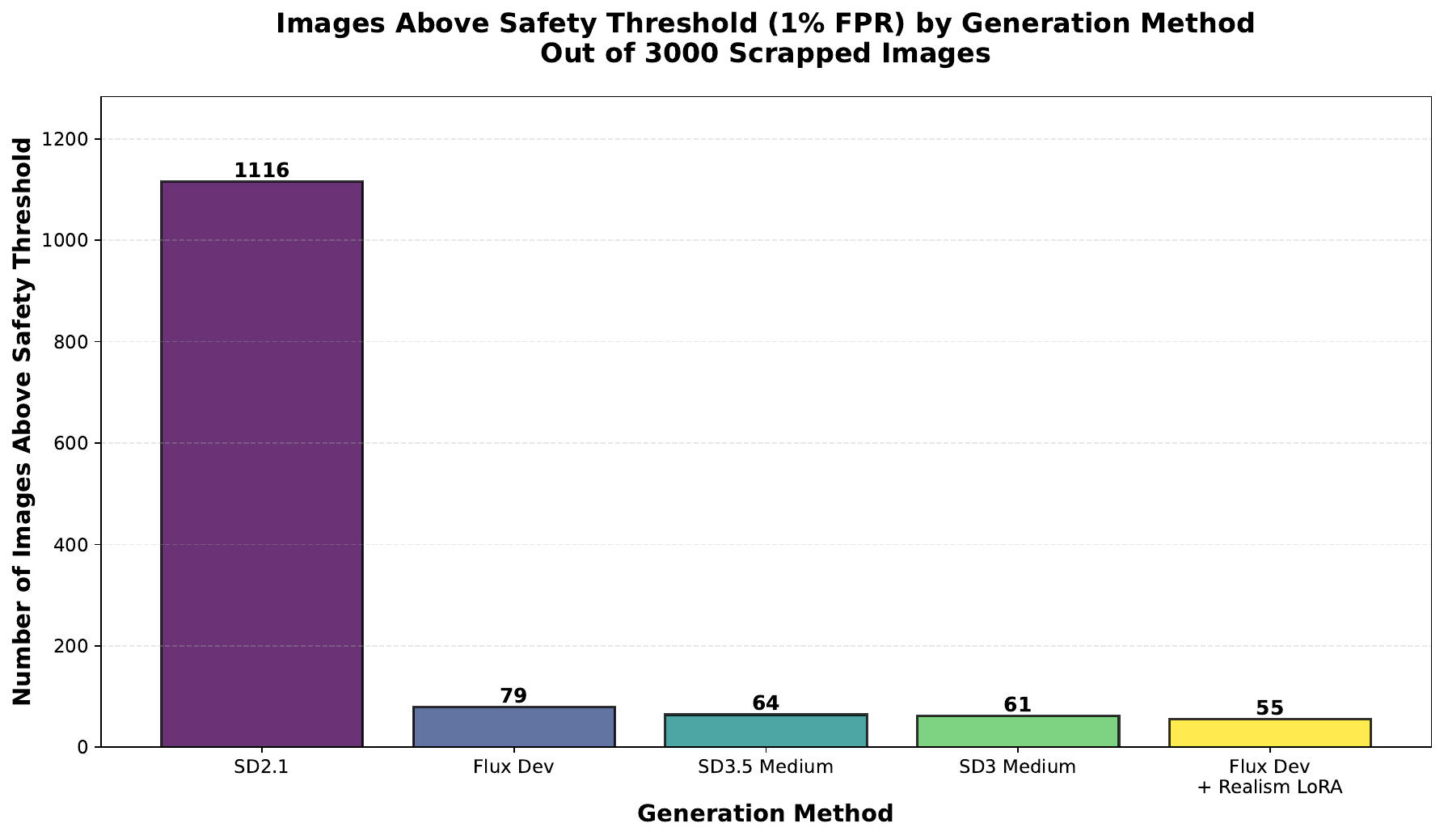}
    \caption{Number of real images (out of 3,000) that exceed the safety threshold, with thresholds calibrated to 1\% FPR for each generative model. Newer models yield only 55–79 images, indicating that modern generators are more capable of closely resynthesizing real-world content, thereby reducing the number of confidently authenticatable samples.}
\vspace{-.5cm}
    \label{fig:scrapped_image}
    
\end{figure}

The motivation for this study is to understand how vulnerable real-world imagery is to resynthesis as generative models become increasingly powerful. If common internet images can be inverted with high fidelity, then their authenticity becomes difficult to establish, since an adversary can plausibly claim they were generated. To examine this risk in practice, we collected approximately 3{,}000 images from Reddit across diverse categories using a range of keywords, and we captioned these images with Salesforce BLIP-2~\cite{blip2}.We are then inverting them using Stable Diffusion 3 (medium). Separately, we calibrate a safety threshold for each generative method Stable Diffusion 2.1, SD3 (medium), SD3.5 (medium), Flux Dev, and Flux Dev with Realism LoRA using distributions of real-inverted and fake-inverted images. This calibration yields method specific safety thresholds: 0.015 for SD2.1, 0.0368 for SD3 (medium), 0.0365 for SD3.5 (medium), 0.035 for Flux Dev, and 0.038 for Flux Dev with Realism LoRA. We then apply each of these thresholds to the same set of SD3 inverted internet images, and count how many images fall above the threshold for each method. As shown in~\Cref{fig:scrapped_image}, SD2.1 marks 1,116 images as authentic, while newer models yield significantly fewer (55–79), reflecting their stronger ability to resynthesize real-world content. These results show a clear progression: newer and larger models can more easily reproduce existing internet imagery. In simple terms, as generative models improve, it becomes increasingly difficult to determine whether an image was captured from reality or synthesized pointing toward an eventual erosion of post-hoc distinguishability between real and generated content.

\subsection{Extension to Video}
To evaluate whether our image-based authentication framework extends to the video domain, we conduct experiments on a subset of 100 videos ( 50 real and 50 fake ) from the Deepfake-Eval-2024 benchmark~\citep{chandra2025deepfakeeval2024multimodalinthewildbenchmark}. Prior work shows that existing detectors suffer approximately 50\% AUC degradation on this benchmark compared to older datasets, highlighting severe generalization failures under real-world distribution shift.

Our method was designed for single images and requires no modification for video. For each video, we sample 8 frames uniformly (one frame per 30-frame interval),then use blip 2~\cite{blip2} to caption then invert and regenerate each frame independently, and compute per-frame authenticity indices $\text{A-index}(f_i, \tilde{f}_i)$ using the same pipeline as our image experiments. The video-level authenticity score is defined as:
\begin{equation}
\text{A-index}_{\text{video}} = \frac{1}{8}\sum_{i=1}^{8} \text{A-index}(f_i, \tilde{f}_i).
\end{equation}

\begin{table}[t]
\centering
\caption{Failure of video-specific deepfake detectors on 50 videos from Deepfake-Eval-2024. All models exhibit poor precision, rendering them unreliable for real-world authentication despite exploiting temporal information.}
\label{tab:video_baseline_failures}
\small
\begin{tabular}{lcccc}
\toprule
\textbf{Model} & \textbf{AUC} & \textbf{Precision} & \textbf{Recall} & \textbf{F1} \\
\midrule
GenConViT~\cite{genconvit} & 0.6154 & \cellcolor{red!15}0.59 & 0.49 & 0.53 \\
FTCN~\cite{ftcn} & \cellcolor{red!15}0.4832 & \cellcolor{red!15}0.50 & 0.64 & 0.40 \\
Styleflow~\cite{styleflow} & 0.5087 & \cellcolor{red!15}0.53 & 0.42 & 0.47 \\
\bottomrule
\end{tabular}
\vspace{-0.5em}
\end{table}

\begin{figure}[ht]
    \centering
    \includegraphics[width=0.9\linewidth]{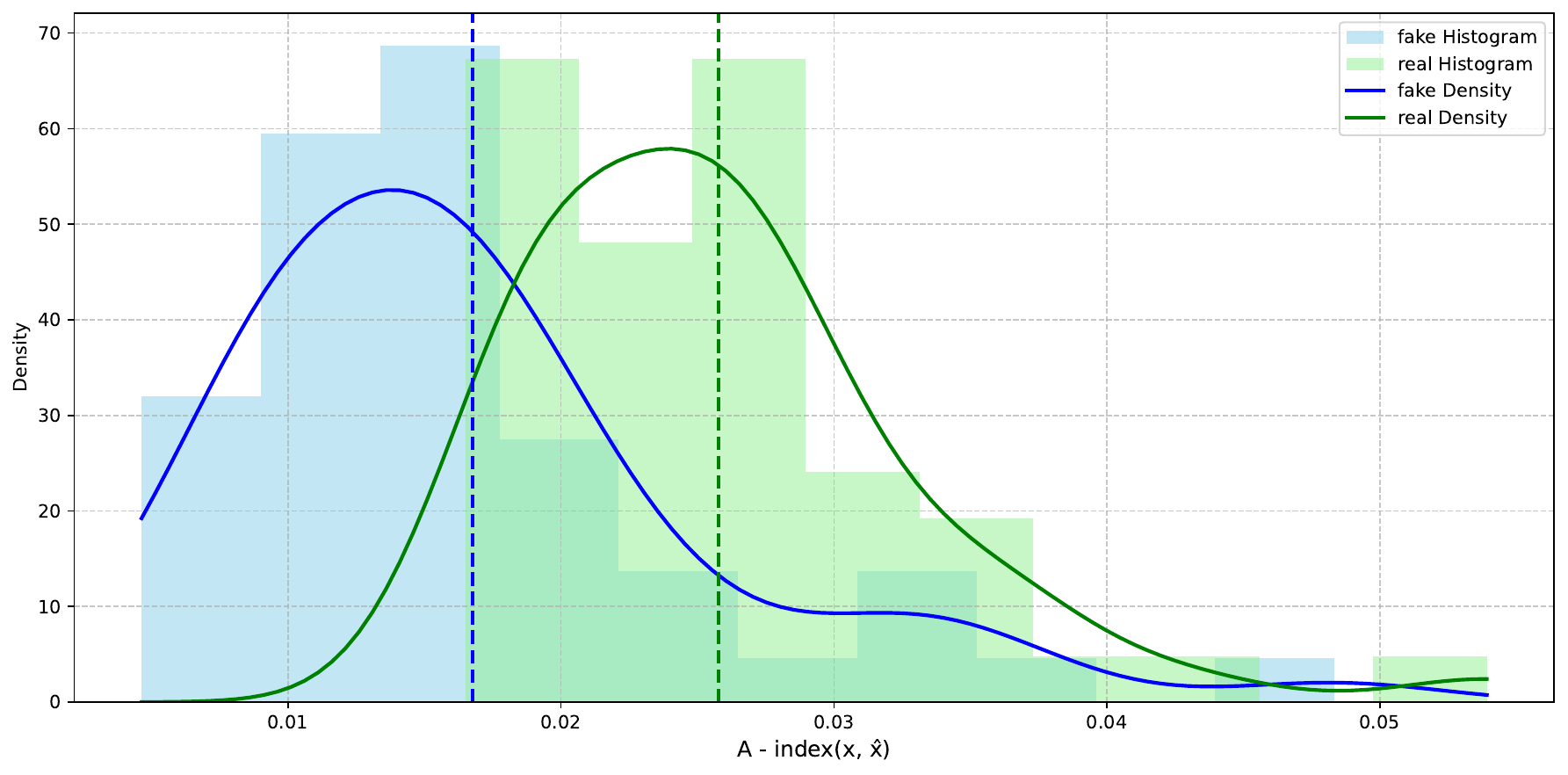}
    \caption{Authenticity Index distribution across real and fake videos}

    \label{fig:video_graph}
    \vspace{-.5cm}
\end{figure}

\tit{Results and Comparison}
\Cref{tab:video_baseline_failures} exposes the precision collapse of purpose-built video deepfake detectors when evaluated on real-world content. Despite being specifically designed to exploit temporal cues and trained on video-specific features, all three baselines fail to achieve reliable authentication capability.GenConViT achieves the highest AUC (0.6154) but maintains only 59\% precision. FTCN performs near-random (AUC 0.4832) and despite achieving 64\% recall through aggressive classification, collapses to 50\% precision, effectively no better than a coin flip. Styleflow shows marginally better performance but still fails with 53\% precision. These results mirror the failures observed with image-based binary classifiers, while our method, as shown in \Cref{fig:video_graph}, exhibits a similar pattern: real videos are harder to invert than fake ones.


\section{Discussion}
As generated images grow more realistic and adversarial, binary detectors fail to generalize and collapse under imperceptible adversarial perturbations. We introduce a calibrated method using an Authenticity Index (A-index) to assess whether content is authentic or plausibly deniable via inversion behavior. By fusing similarity metrics into a single score, our approach enables strict, attack-resistant thresholds. Tested on images, videos, and internet scrapped images, it yields high precision, low recall, and abstains in uncertain cases. Stronger generators, especially LoRA-based, further shrink verifiable content. Our method generalizes to video and supports a shift toward risk-aware, abstention-based detection.

{
    \small
    \bibliographystyle{ieeenat_fullname}
    \bibliography{main}
}

\newpage
\maketitlesupplementary

\setcounter{section}{0}
\setcounter{figure}{0}    
\setcounter{table}{0}  
\renewcommand{\thetable}{\Alph{table}}
\renewcommand{\thefigure}{\Alph{figure}}
\renewcommand{\thesection}{\Alph{section}}

\section{Evaluation metric}
\subsection{Similarity Metrics Formulation}
In this section, we provide the mathematical formulations for all similarity metrics used in computing the Authenticity Index.
\tit{Peak Signal-to-Noise Ratio (PSNR)}
PSNR measures the pixel-level fidelity between the original image $\mathbf{x}$ and its reconstruction $\tilde{\mathbf{x}}$:

\begin{equation}
\text{PSNR}(\mathbf{x}, \tilde{\mathbf{x}}) = 10 \cdot \log_{10}\left(\frac{\text{MAX}_I^2}{\text{MSE}(\mathbf{x}, \tilde{\mathbf{x}})}\right)
\end{equation}
where $\text{MSE}(\mathbf{x}, \tilde{\mathbf{x}}) = \frac{1}{N} \sum_{i=1}^{N} (x_i - \tilde{x}_i)^2$, $\text{MAX}_I$ is the maximum possible pixel value (255 for 8-bit images), and $N$ is the total number of pixels. PSNR is measured in decibels (dB), with higher values indicating better reconstruction quality.
\tit{Structural Similarity Index (SSIM)} SSIM~\cite{ssim} evaluates structural information preservation between images:
\begin{equation}
\text{SSIM}(I, \hat{I}) = \frac{(2\mu_I \mu_{\hat{I}} + C_1)(2\sigma_{I\hat{I}} + C_2)}{(\mu_I^2 + \mu_{\hat{I}}^2 + C_1)(\sigma_I^2 + \sigma_{\hat{I}}^2 + C_2)},
\end{equation}
where $\mu$, $\sigma^2$, and $\sigma_{I\hat{I}}$ are the local means, variances, and cross-covariance of $I$ and $\hat{I}$, and $C_1$, $C_2$ are stabilizing constants.
\tit{Learned Perceptual Image Patch Similarity (LPIPS)} LPIPS~\cite{lpips} computes perceptual distance using deep features extracted from a pretrained network:
\begin{equation}
\text{LPIPS}(\mathbf{x}, \tilde{\mathbf{x}}) = \sum_{\ell} \frac{1}{H_\ell W_\ell} \sum_{h,w} \|\mathbf{w}_\ell \odot (\phi_\ell(\mathbf{x})_{h,w} - \phi_\ell(\tilde{\mathbf{x}})_{h,w})\|_2^2
\end{equation}
where $\phi_\ell$ represents features from layer $\ell$ of a pretrained network, $\mathbf{w}_\ell$ are learned weights, and $H_\ell, W_\ell$ are spatial dimensions at layer $\ell$. Since LPIPS measures distance, we use $(1 - \text{LPIPS})$ in our formulation to align with other similarity metrics.
\tit{CLIP Similarity}
CLIP~\cite{clip} measures semantic consistency in the joint vision-language embedding space:

\begin{equation}
\text{CLIP}(\mathbf{x}, \tilde{\mathbf{x}}) = \frac{E(\mathbf{x}) \cdot E(\tilde{\mathbf{x}})}{\|E(\mathbf{x})\| \cdot \|E(\tilde{\mathbf{x}})\|}
\end{equation}
where $E(\cdot)$ is the CLIP vision encoder, and the similarity is computed as the cosine similarity between the normalized embeddings.
\subsection{Authenticity Index Calibration}
\tit{Weight Optimization for the composite similarity score}
Given our Similarity score:
\begin{equation}
\begin{split}
s(x,\tilde{x}) =\;& \alpha_1 \cdot \text{PSNR}(x,\tilde{x}) + \alpha_2 \cdot \text{SSIM}(x,\tilde{x}) \\
&+ \alpha_3 \cdot (1 - \text{LPIPS}(x,\tilde{x})) + \alpha_4 \cdot \text{CLIP}(x,\tilde{x}).
\end{split}
\end{equation}
and our Authenticity index:
\begin{equation}
\text{A-index}(\mathbf{x}, \tilde{\mathbf{x}}) = \frac{\exp(-\sigma \cdot s(\mathbf{x}, \tilde{\mathbf{x}}))}{1 + \exp(-\sigma \cdot s(\mathbf{x}, \tilde{\mathbf{x}}))}
\end{equation}
we aim to optimize the weights $\alpha_1, \alpha_2, \alpha_3, \alpha_4$ via using Differential Evolution to maximize separation between real and fake image distributions. The objective function minimizes the overlap between distributions:

\begin{equation}
\text{minimize} \quad \text{overlap}(\mathcal{D}_{\text{real}}, \mathcal{D}_{\text{fake}}) = \int \min(p_{\text{real}}(s), p_{\text{fake}}(s)) \, ds
\end{equation}

\tit{Differential Evolution Parameters}
\begin{itemize}
    \item Population size: 20
    \item Mutation factor ($F$): 0.6
    \item Crossover probability ($CR$): 0.7
    \item Maximum iterations: 300
    \item Bounds: $\alpha_i \in [-10.0, 10.0]$ for $i = 1,2,3,4$
    \item Convergence tolerance: $1 \times 10^{-10}$
\end{itemize}

\tit{Optimized Weights}
\begin{itemize}
    \item $\alpha_1$ (PSNR): -0.0181
    \item $\alpha_2$ (SSIM): 1.380
    \item $\alpha_3$ (1-LPIPS): -4.058
    \item $\alpha_4$ (CLIP): 8.066
    \item $\sigma$ (sigmoid scale): 0.9
\end{itemize}

\section{Inversion Hyperparameters}
We employ Rectified Flow Inversion~\cite{rf_inversion} for all experiments. The key hyperparameters are described below.
\tit{General Inversion Parameters}
\begin{table}[H]
\centering
\resizebox{\columnwidth}{!}{%
\begin{tabular}{ccc}
\toprule
\textbf{Parameter} & \textbf{Value} & \textbf{Description} \\
\midrule
Number of inference steps & 28 & Denoising timesteps \\
Guidance scale & 3.5 & Classifier-free guidance strength \\
Scheduler & FlowMatchEulerDiscrete & Sampling scheduler used during inversion \\
DDIM $\eta$ base & 0.95 & Base eta for interpolated denoise \\
Eta trend & constant & Trend of eta values (e.g., constant, linear) \\
Eta start step & 0 & Step where eta schedule starts \\
Eta end step & 9 & Step where eta schedule ends \\
Gamma & 0.5 & Gamma parameter for rectified inversion \\
Seed & 42 & Random seed for reproducibility \\
Data type & float16 & Precision for model computations \\
\bottomrule
\end{tabular}
}
\caption{General inversion hyperparameters used across all experiments.}
\end{table}

\section{Zero-Shot Detection Graphs for All Baseline Methods}

We report the prediction distributions of all baseline detectors under a zero-shot setting.These detectors include UFD~\cite{ufd}, FreqNet~\cite{fakedetect6}, NPR~\cite{npr}, FatFormer~\cite{fatformer}, D3~\cite{d3}, and C2P-CLIP~\cite{c2pclip}. The evaluation was conducted on test samples generated by models unseen during training, namely Stable Diffusion 2.1, Stable Diffusion XL, Stable Diffusion 3 (medium), Stable Diffusion 3.5 (medium), DALL·E 3, and Midjourney 6.

\begin{figure*}[t]
    \centering
    \begin{subfigure}[b]{0.5\linewidth}
    \includegraphics[width=\linewidth]{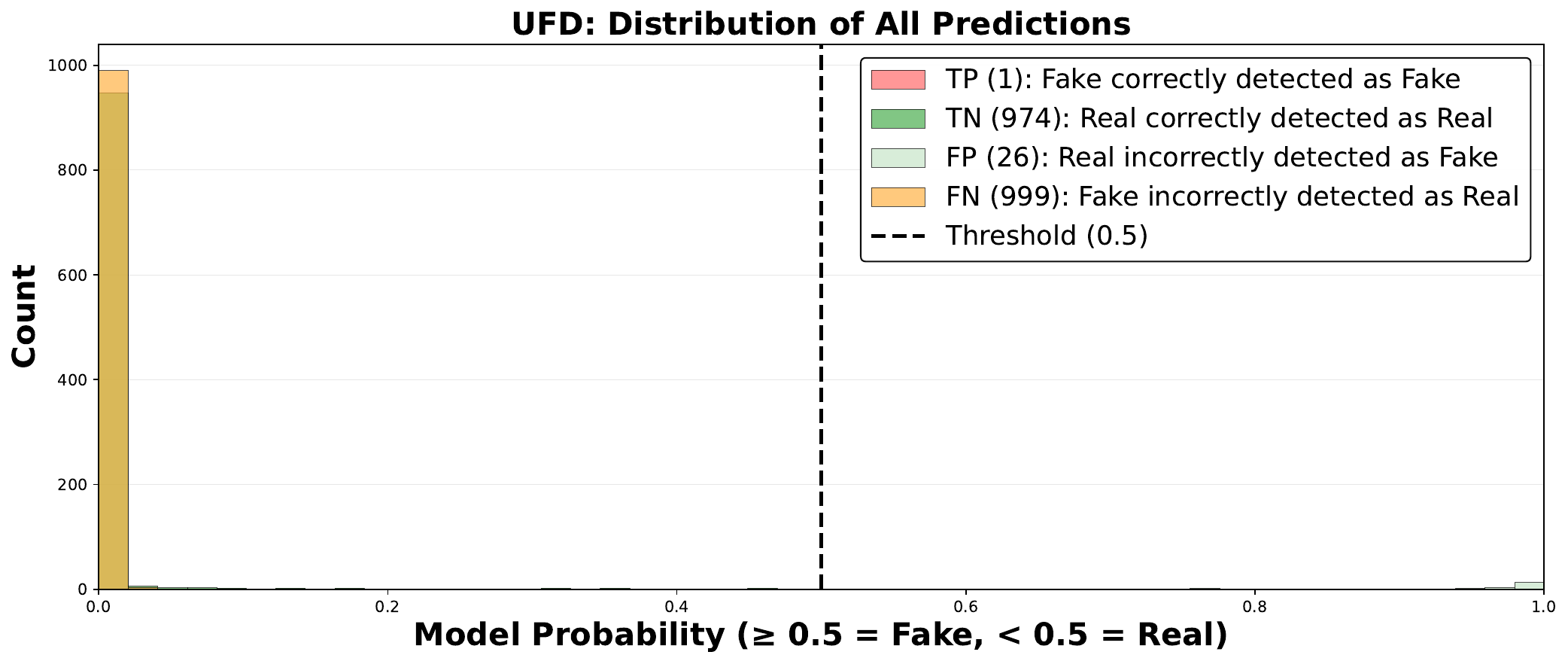}
        \caption{UFD}
    \end{subfigure}
    \hfill
    \begin{subfigure}[b]{0.48\linewidth}
        \includegraphics[width=\linewidth]{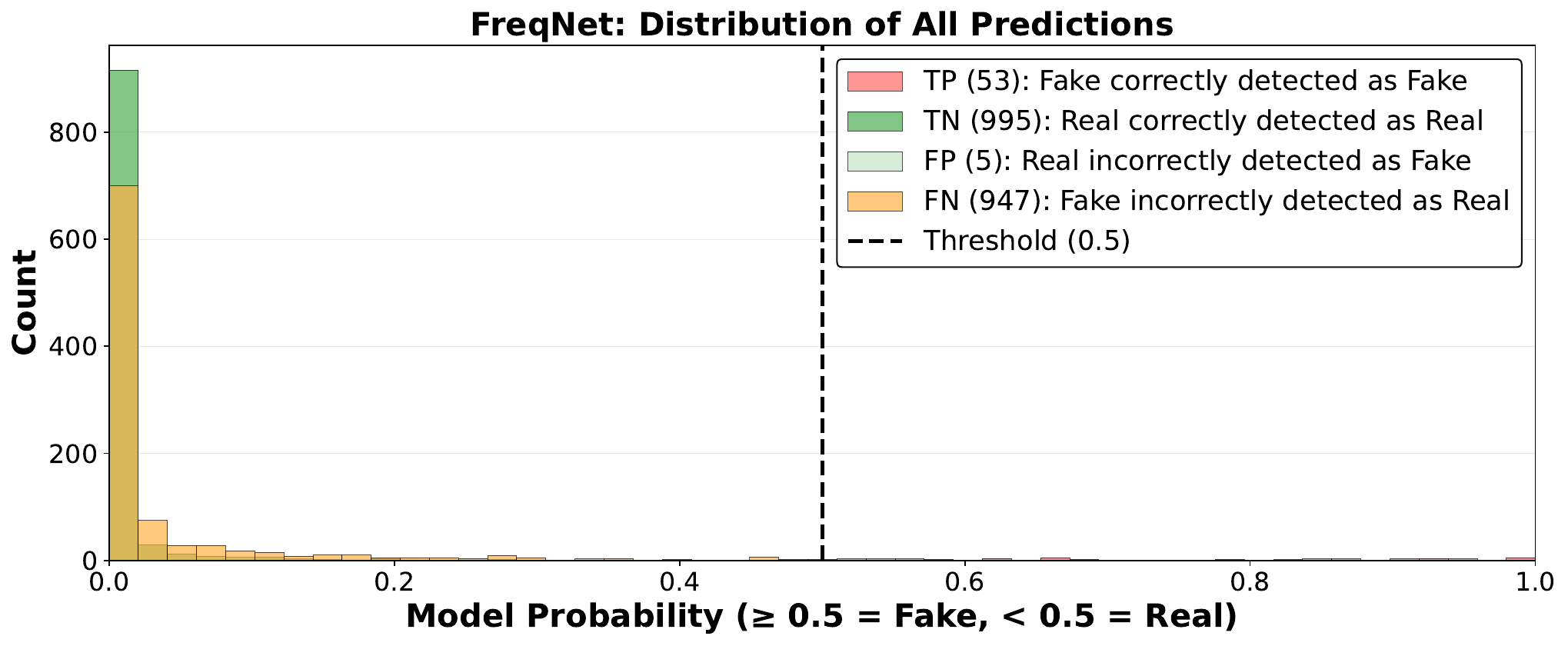}
        \caption{FreqNet}
    \end{subfigure}

    \vspace{0.1cm}

    \begin{subfigure}[b]{0.48\linewidth}
        \includegraphics[width=\linewidth]{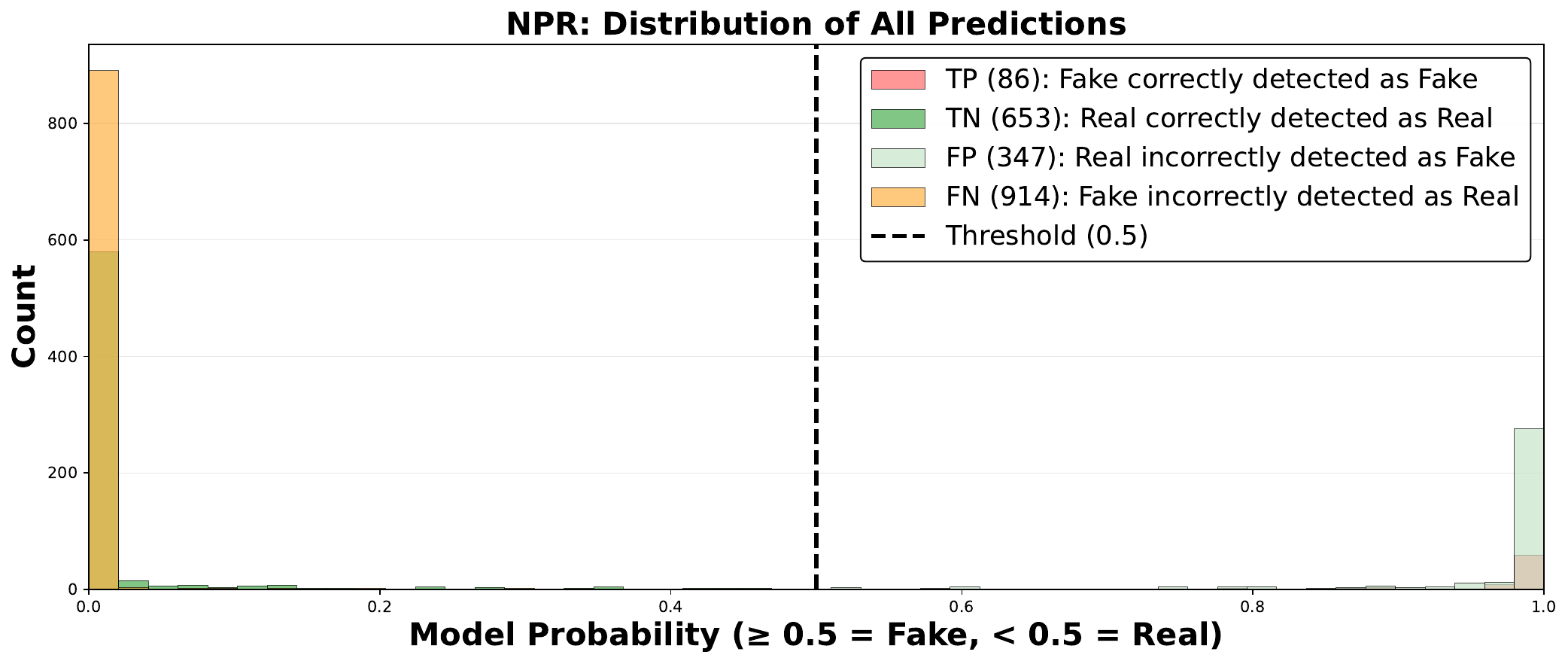}
        \caption{NPR}
    \end{subfigure}
    \hfill
    \begin{subfigure}[b]{0.48\linewidth}
        \includegraphics[width=\linewidth]{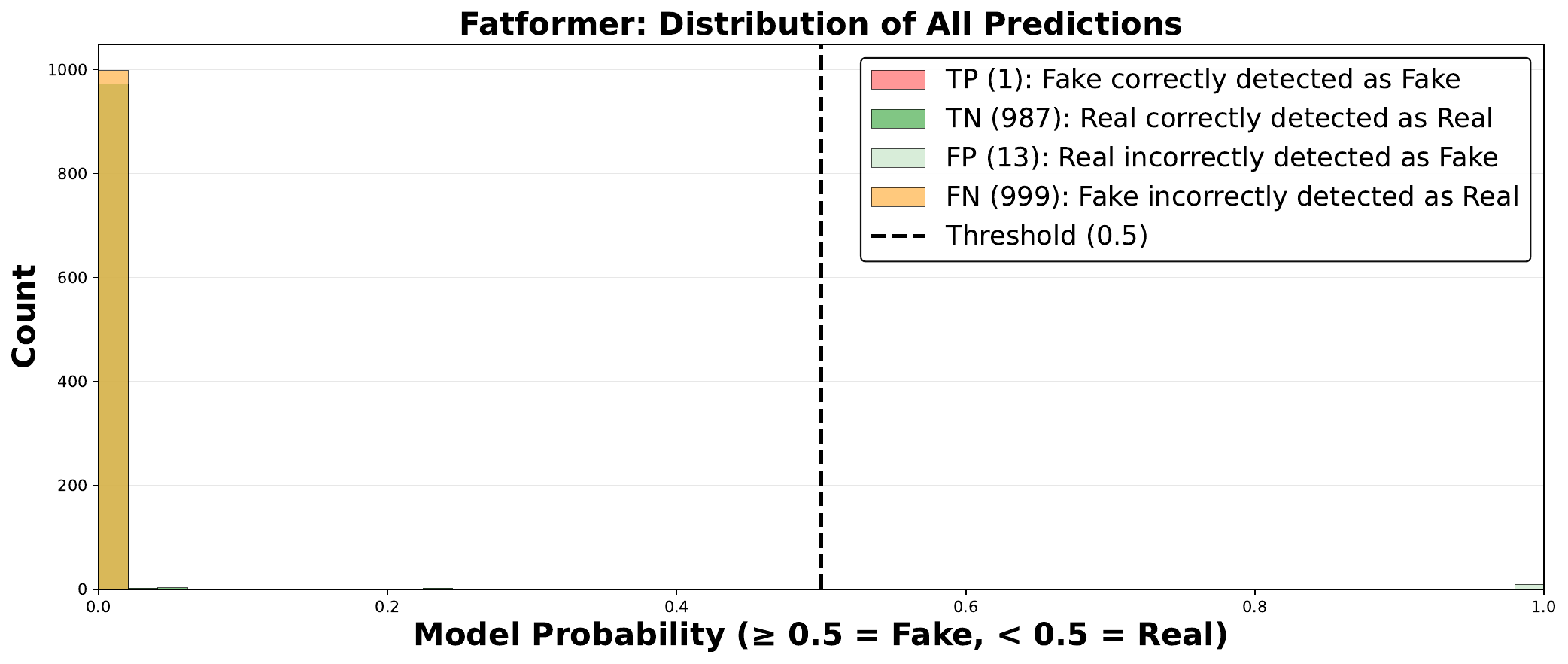}
        \caption{FatFormer}
    \end{subfigure}

    \vspace{0.1cm}

    \begin{subfigure}[b]{0.48\linewidth}
        \includegraphics[width=\linewidth]{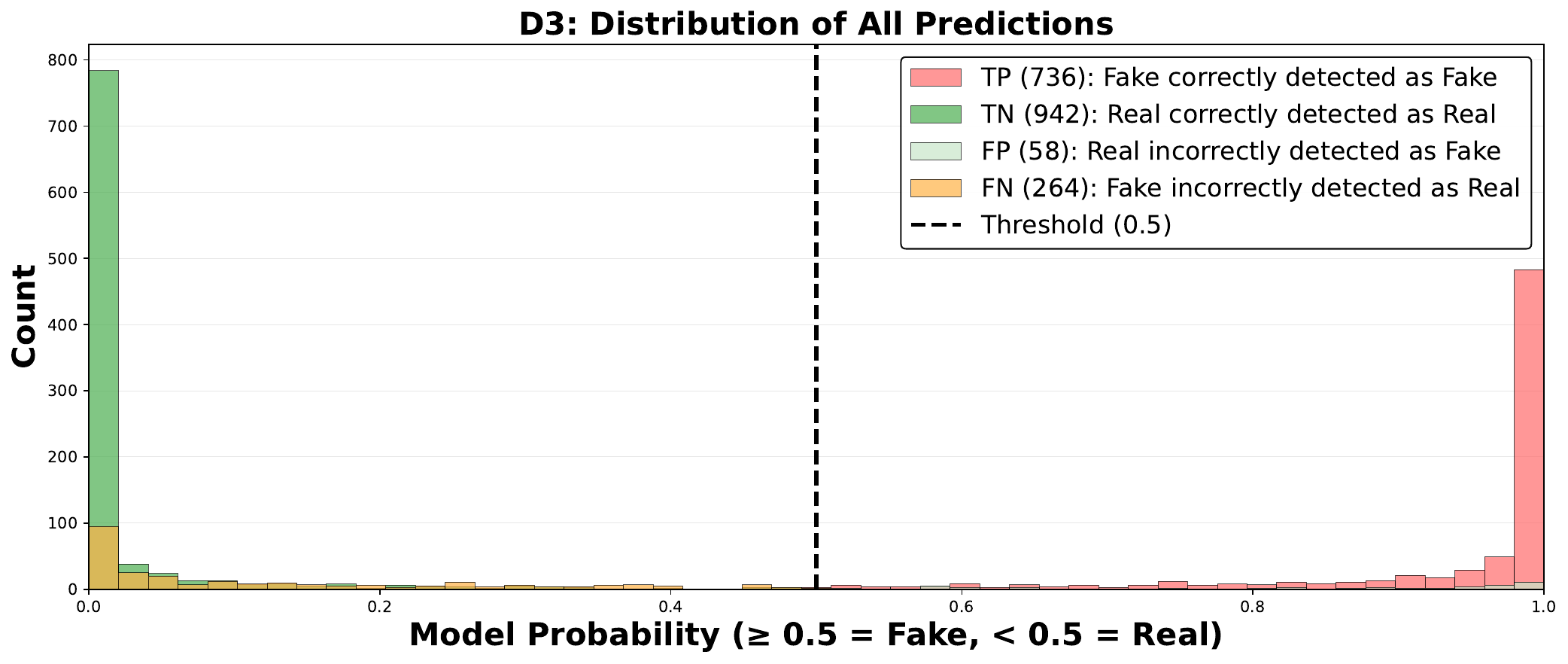}
        \caption{D3}
    \end{subfigure}

    \caption{Zero-shot prediction score distributions for each baseline method. Each subplot shows the histogram of classifier outputs for real and fake images. Vertical dashed lines indicate classification thresholds. most methods struggle to generalize to unseen generators, misclassifying a significant number of fake images as real.}
    \label{fig:zero_shot_all}
\end{figure*}

Our observations indicate that most methods exhibit substantial prediction score overlap between real and fake classes. As shown in Figure \ref{fig:zero_shot_all}, models like UFD, FreqNet, and NPR tend to overfit to familiar artifacts and thus default to high-confidence predictions of “real” even when presented with synthetic samples from unseen generators. Although D3 performs slightly better, it still suffers from notable misclassification rates. These plots underline the generalization weakness of current binary detectors and reinforce the motivation behind adopting a calibrated authenticity framework.

\section{PGD Attack Graphs for All Baseline Methods}

To further assess robustness, we visualize how the prediction distributions of five baseline detectors change under adversarial perturbation. Specifically, we perform projected gradient descent (PGD) attacks~\cite{pgd} with $\ell_\infty$-norm bounded perturbations ($\epsilon = 8/255$), targeting both real and fake inputs. The objective is to push predictions across the decision boundary using imperceptible noise.

\begin{figure*}[!t]
    \centering
    \begin{subfigure}[b]{0.45\linewidth}
        \includegraphics[width=\linewidth]{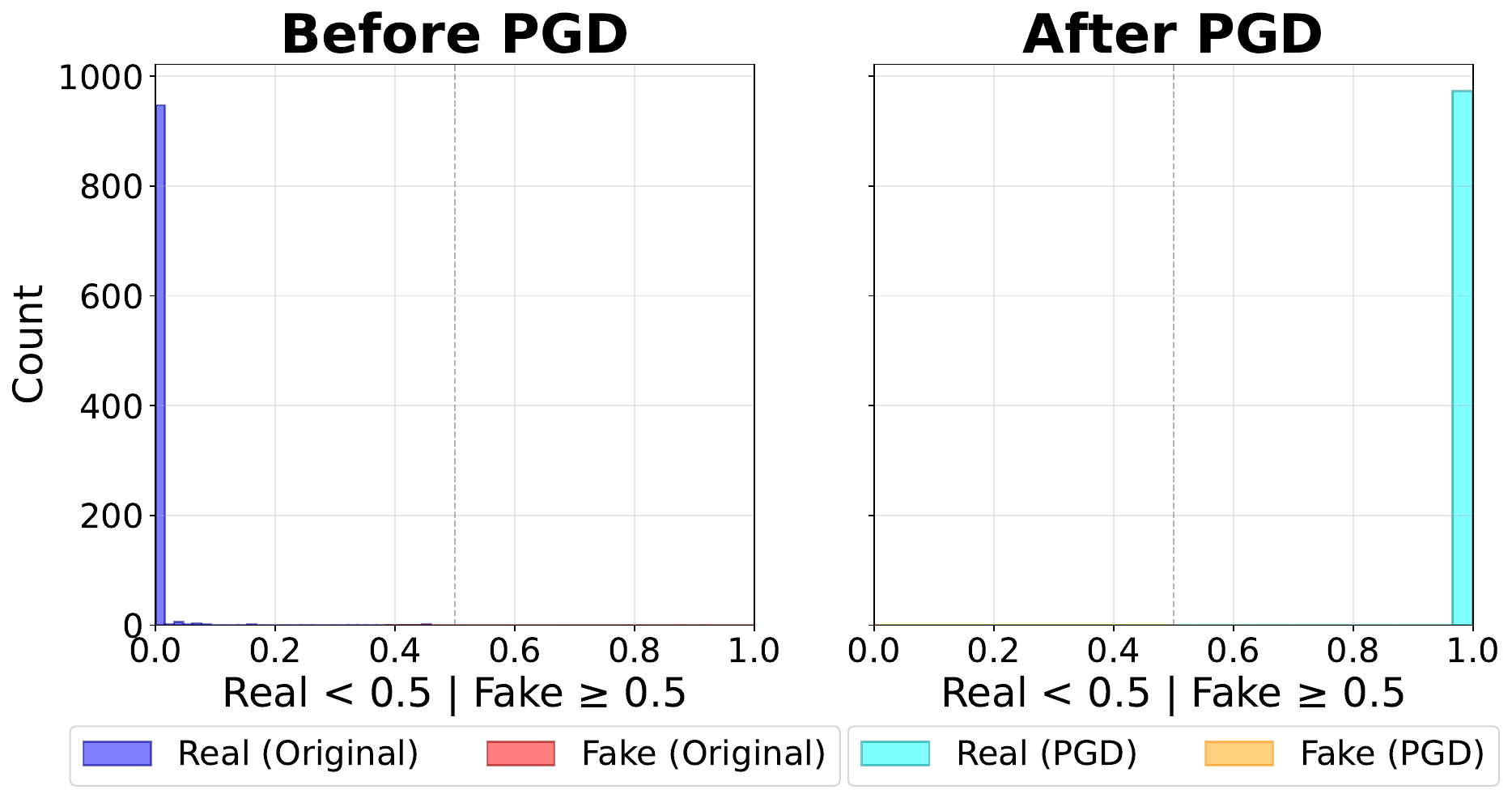}
        \caption{UFD}
    \end{subfigure}
    \hfill
    \begin{subfigure}[b]{0.45\linewidth}
        \includegraphics[width=\linewidth]{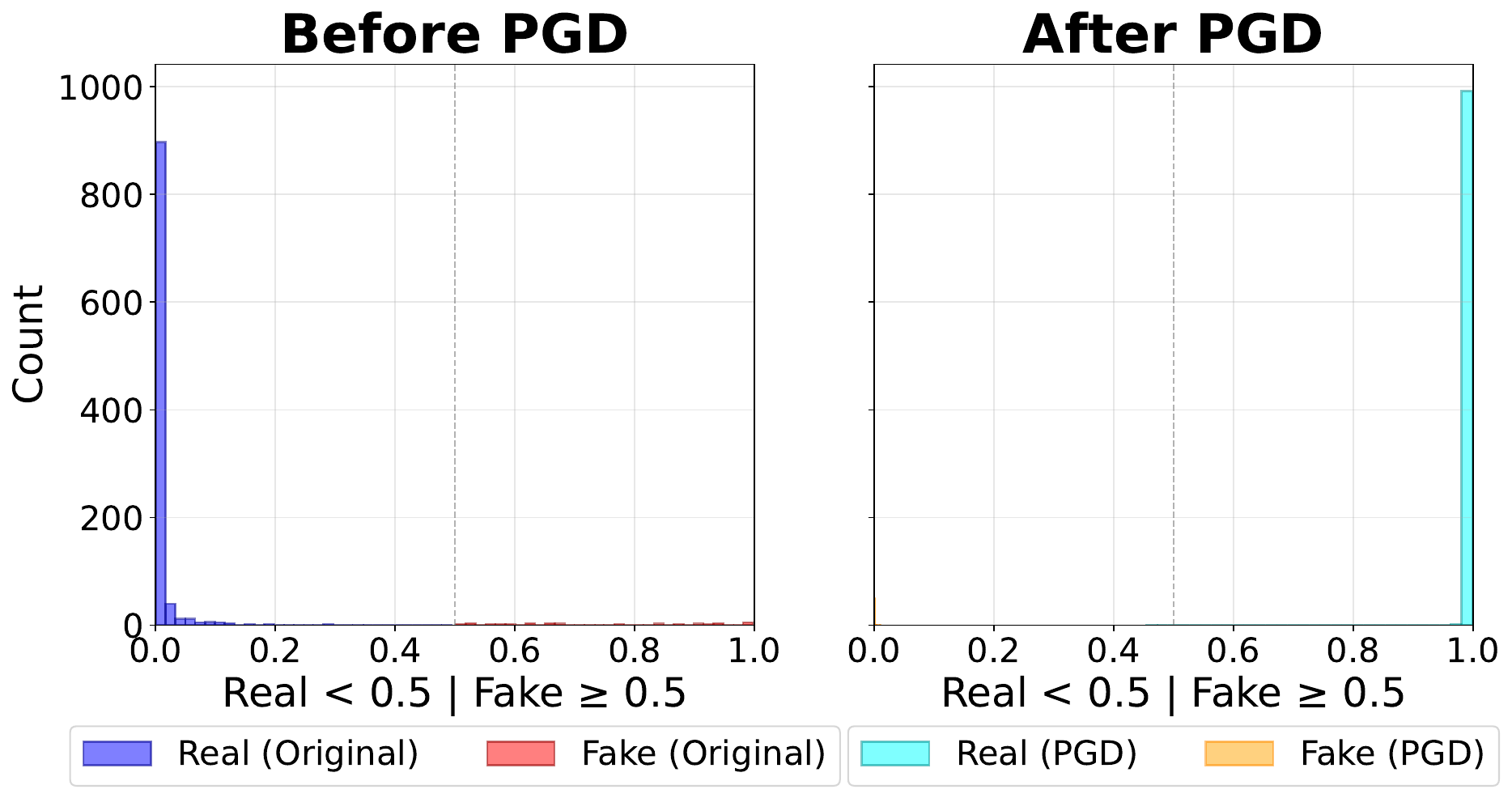}
        \caption{FreqNet}
    \end{subfigure}

    \vspace{0.1cm}

    \begin{subfigure}[b]{0.48\linewidth}
        \includegraphics[width=\linewidth]{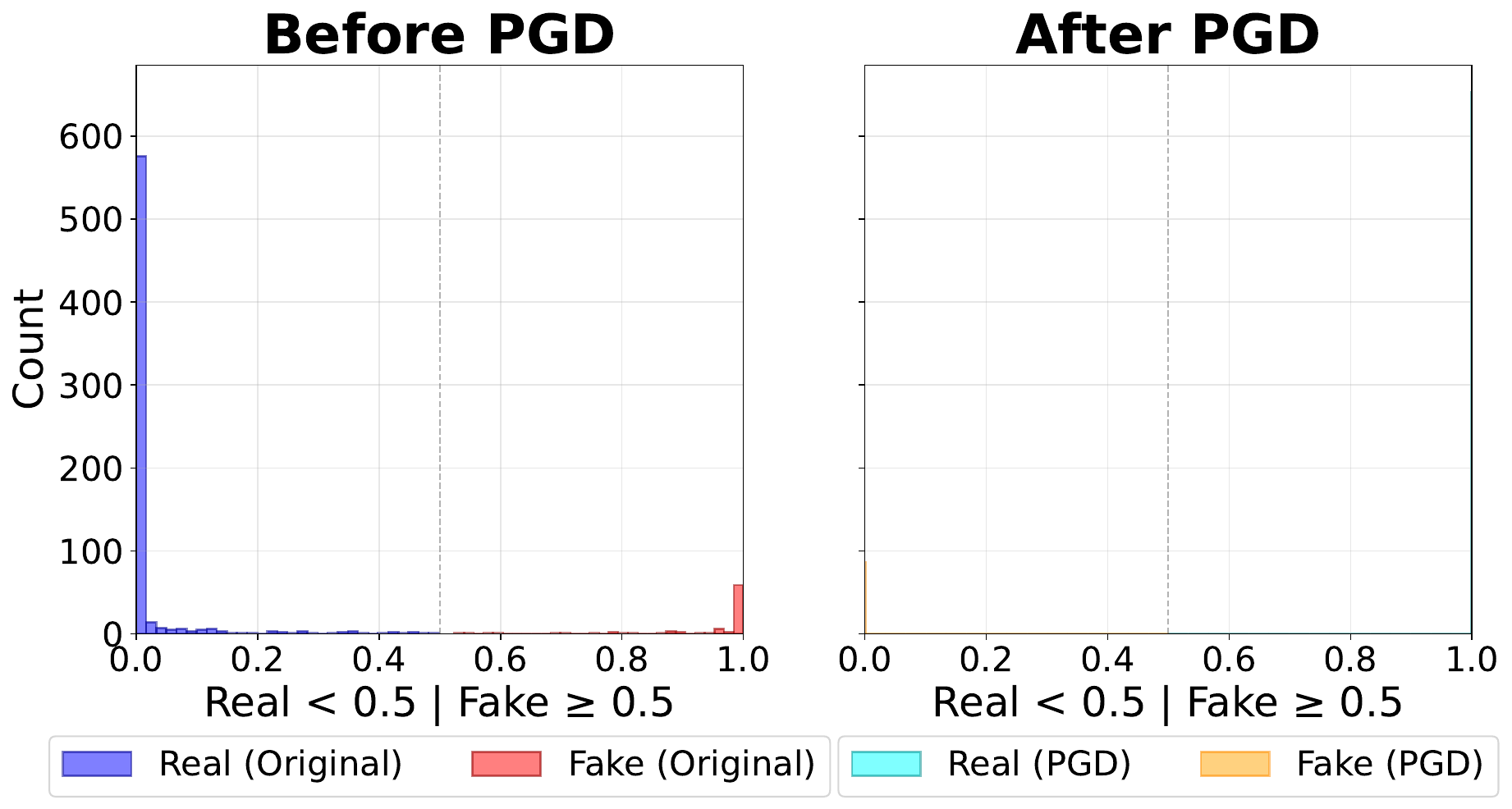}
        \caption{NPR}
        \label{fig:pgd_attacks_c}
    \end{subfigure}
    \hfill
    \begin{subfigure}[b]{0.48\linewidth}
        \includegraphics[width=\linewidth]{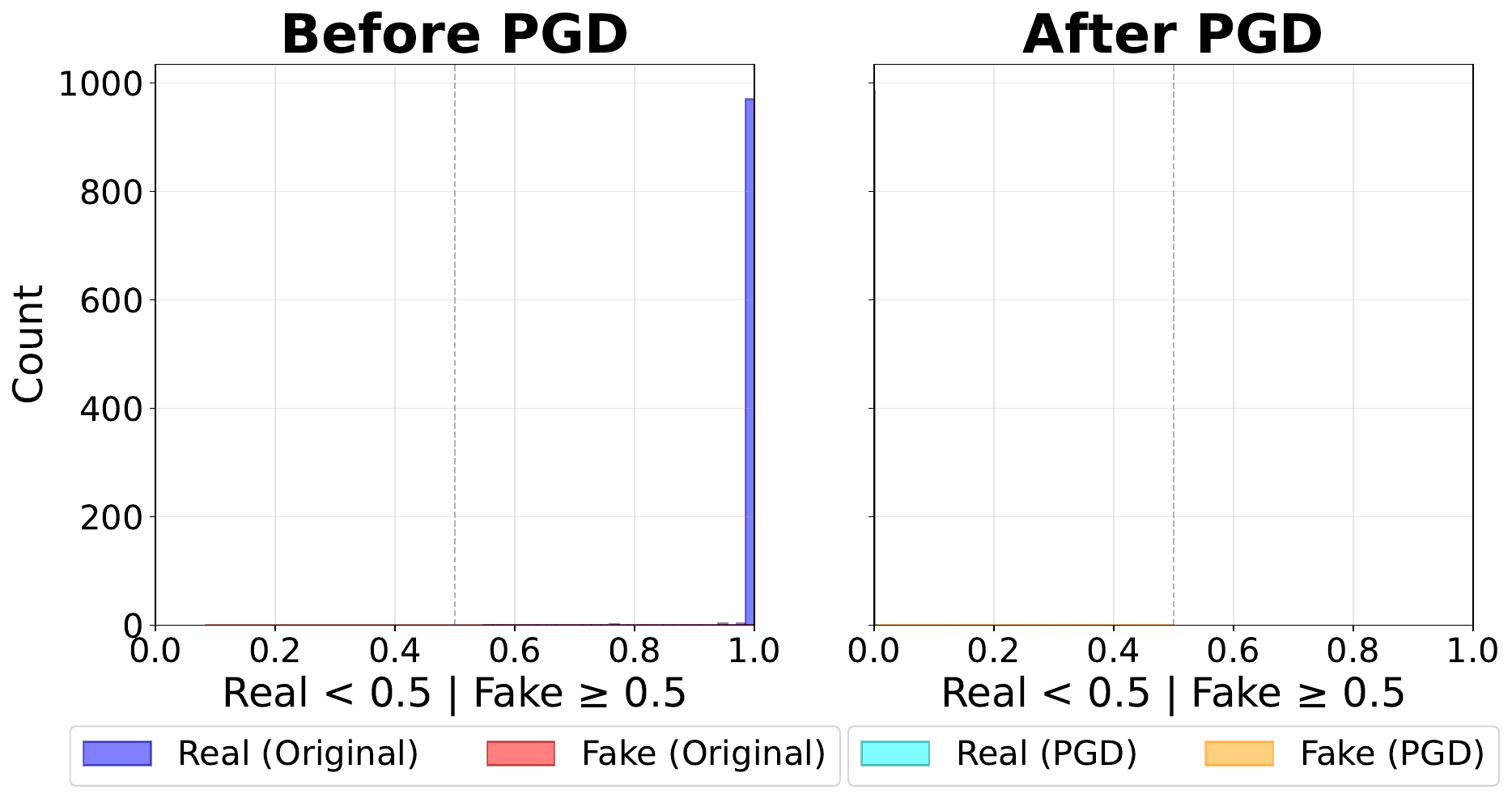}
        \caption{FatFormer}
        \label{fig:pgd_attacks_d}
    \end{subfigure}

    \vspace{0.1cm}

    \begin{subfigure}[b]{0.48\linewidth}
        \includegraphics[width=\linewidth]{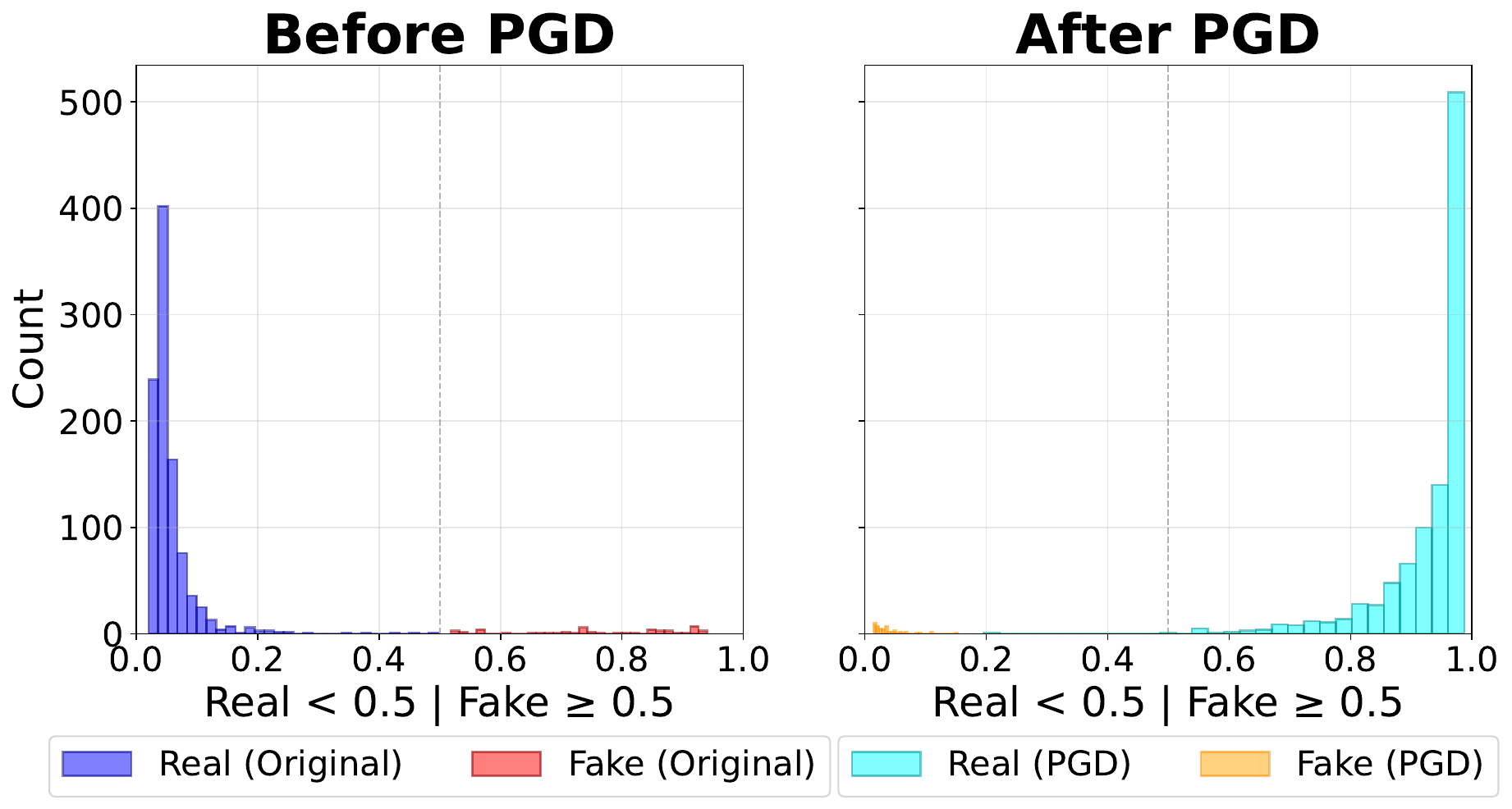}
        \caption{C2P-CLIP}
    \end{subfigure}

    \caption{Prediction score distributions before and after PGD attack for each baseline method (excluding D3). Each plot shows how adversarial perturbation erodes separability between real and fake inputs, often collapsing confidence entirely.}
    \label{fig:pgd_attacks}
\end{figure*}

The results show a complete collapse in performance across all methods. Models like UFD~\cite{fakedetect3}, FreqNet~\cite{fakedetect6}, and NPR~\cite{npr} fail entirely, with prediction distributions collapsing into indistinguishability between real and fake classes. FatFormer~\cite{fatformer}
and C2P-CLIP~\cite{c2pclip} also misclassify nearly all samples post-attack. These visualizations in Figure~\ref{fig:pgd_attacks} highlight the fragility of binary detectors and the lack of graceful degradation under adversarial pressure (a phenomenon that is particularly pronounced in the cases of NPR and FatFormer). In contrast, our proposed method maintains calibrated separation and abstains when confidence is insufficient, as demonstrated in the main paper.

To illustrate the imperceptibility of the attack, Figure~\ref{fig:pgd_visual} shows a real image before and after PGD perturbation, along with the normalized perturbation itself. Despite the negligible visual difference, the prediction shifts significantly post-attack, causing traditional detectors to fail.

\begin{figure*}[!t]
    \centering
    \begin{subfigure}[b]{0.45\linewidth}
        \includegraphics[width=\linewidth]{diagrams/a_brown_and_white_dog_is_sitting_in_some_grass_and_a_red_and_white_fire_hydrant_original.pdf}
        \caption{Original Image}
    \end{subfigure}
    \hfill
    \begin{subfigure}[b]{0.45\linewidth}
        \includegraphics[width=\linewidth]{diagrams/a_brown_and_white_dog_is_sitting_in_some_grass_and_a_red_and_white_fire_hydrant_adversarial.pdf}
        \caption{Adversarially Perturbed Image}
    \end{subfigure}

    \caption{Visual comparison between the original image and its PGD-perturbed version. The perturbation is imperceptible to the human eye, yet it significantly alters detector predictions.}
    \label{fig:pgd_visual}
\end{figure*}

\begin{figure*}[!t]
    \centering
    \includegraphics[width=\linewidth]{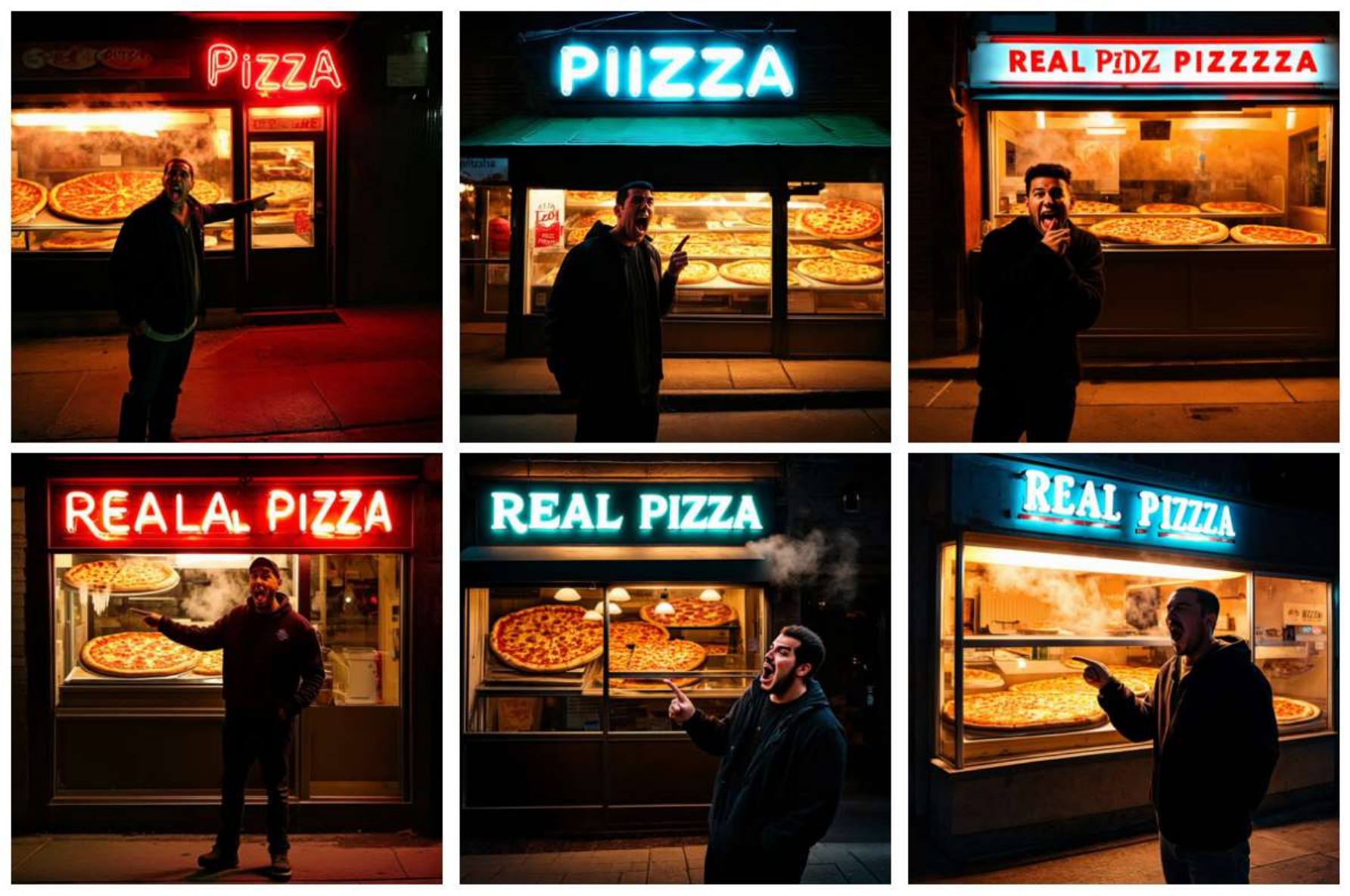}
    \caption{Six image samples generated from the same prompt (\Cref{quote:prompt_hungry_man_real_pizza}), with different random seeds. All were generated under the medium-resource attacker's sampling budget. These highlight the diversity in candidate outputs that the attacker can choose from.}
    \label{fig:medium_attacker_variants}
\end{figure*}

\section{Examples of Hard-to-Invert and Easy-to-Invert Images}

This section presents qualitative examples of images that fall on opposite ends of the Authenticity Index spectrum. We distinguish between “hard-to-invert” images, which yield low similarity under inversion and thus achieve high A-index scores, and “easy-to-invert” images, which reconstruct with high perceptual fidelity and result in low A-index values.

Images that are hard to invert typically contain fine-grained texture, complex scenes, occlusions, or motion blur factors that challenge the inversion pipeline. In contrast, easy-to-invert images often include centered objects with minimal background clutter, strong lighting, or symmetry, all of which tend to align well with the priors of modern generative models.

\begin{figure*}[h]
    \centering

    \begin{subfigure}[b]{0.23\linewidth}
        \includegraphics[width=\linewidth]{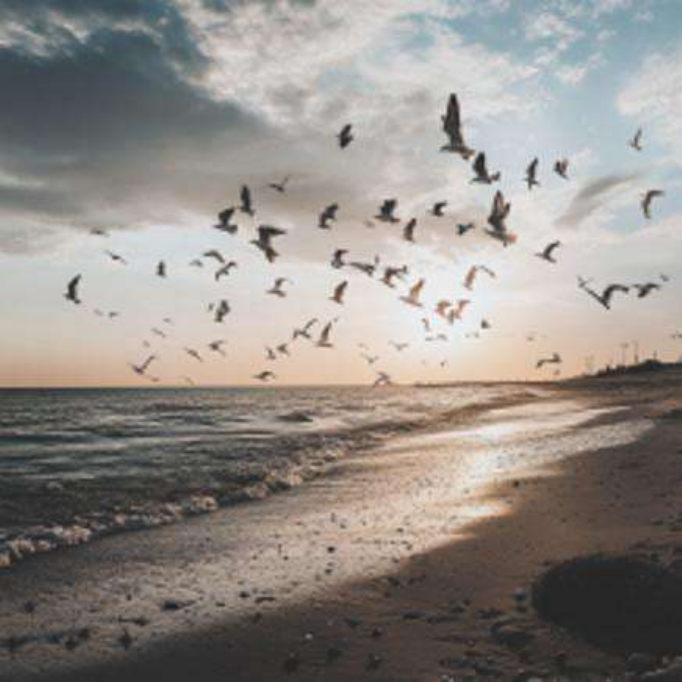}
        \caption*{\footnotesize (a) Original (Easy to invert)}
    \end{subfigure}
    \hfill
    \begin{subfigure}[b]{0.23\linewidth}
        \includegraphics[width=\linewidth]{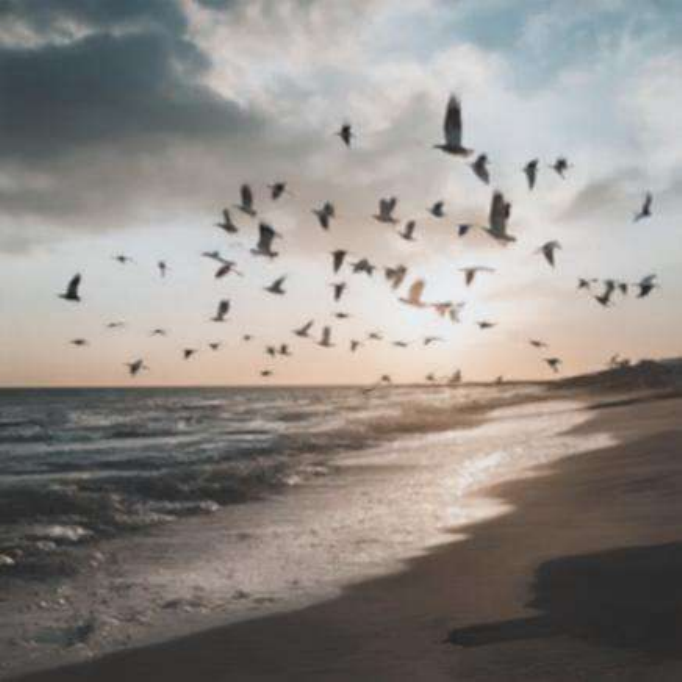}
        \caption*{\footnotesize (b) Inverted (Easy to invert)}
    \end{subfigure}
    \hfill
    \begin{subfigure}[b]{0.23\linewidth}
        \includegraphics[width=\linewidth]{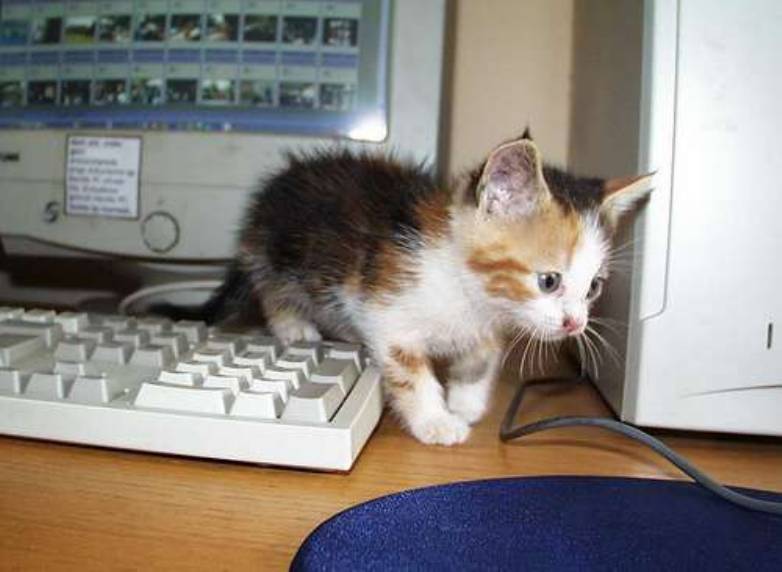}
        \caption*{\footnotesize (c) Original (Easy to invert)}
    \end{subfigure}
    \hfill
    \begin{subfigure}[b]{0.23\linewidth}
        \includegraphics[width=\linewidth]{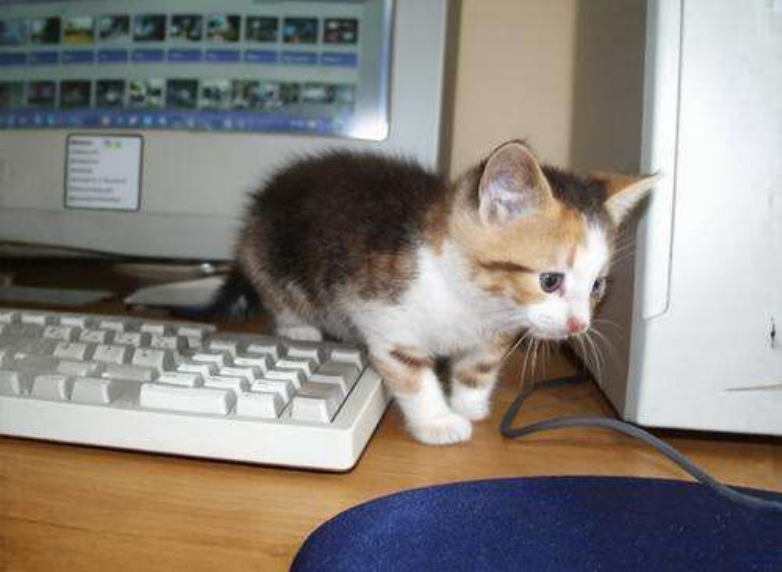}
        \caption*{\footnotesize (d) Inverted (Easy to invert)}
    \end{subfigure}

    \vspace{0.3cm}

    \begin{subfigure}[b]{0.23\linewidth}
        \includegraphics[width=\linewidth]{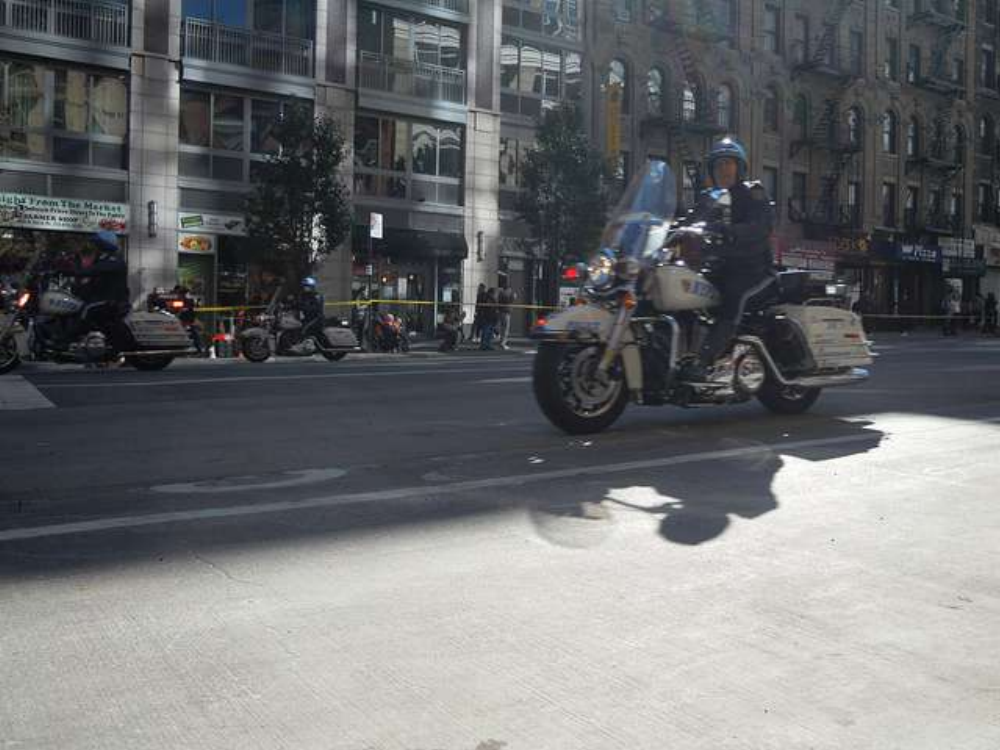}
        \caption*{\footnotesize (e) Original (Hard to invert)}
    \end{subfigure}
    \hfill
    \begin{subfigure}[b]{0.23\linewidth}
        \includegraphics[width=\linewidth]{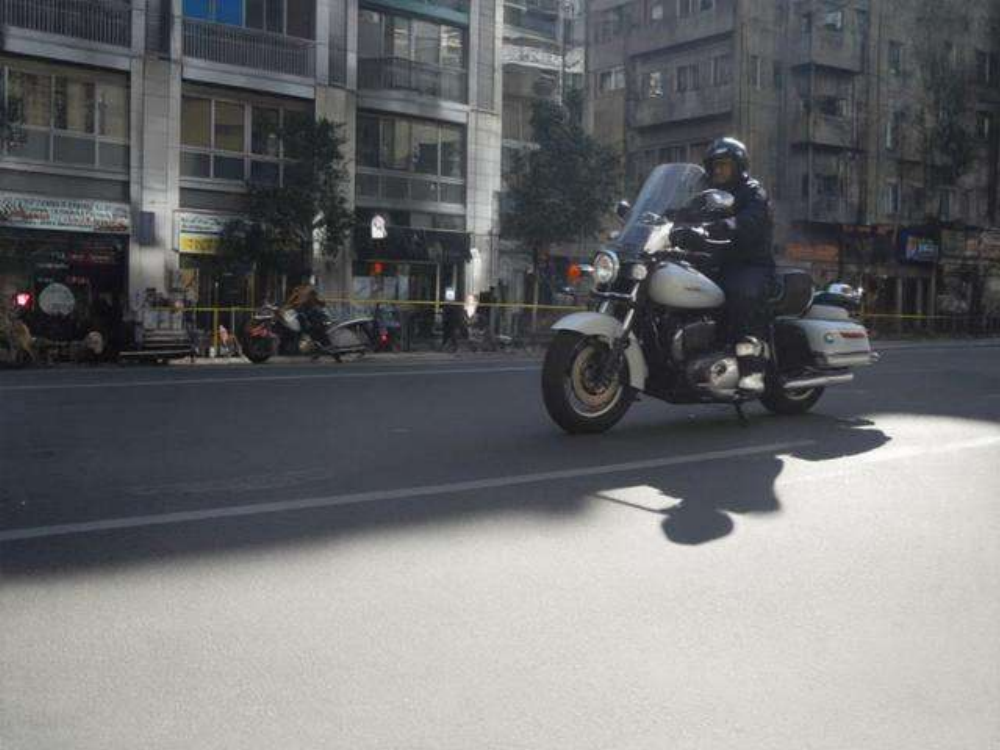}
        \caption*{\footnotesize (f) Inverted (Hard to invert)}
    \end{subfigure}
    \hfill
    \begin{subfigure}[b]{0.23\linewidth}
        \includegraphics[width=\linewidth]{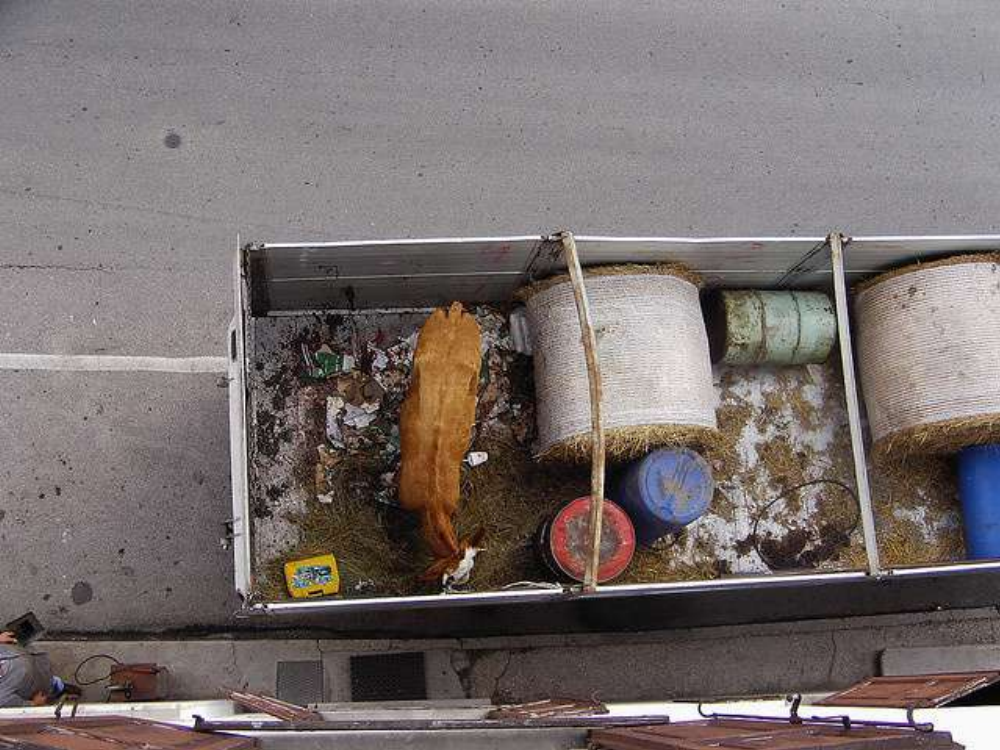}
        \caption*{\footnotesize (g) Original (Hard to invert)}
    \end{subfigure}
    \hfill
    \begin{subfigure}[b]{0.23\linewidth}
        \includegraphics[width=\linewidth]{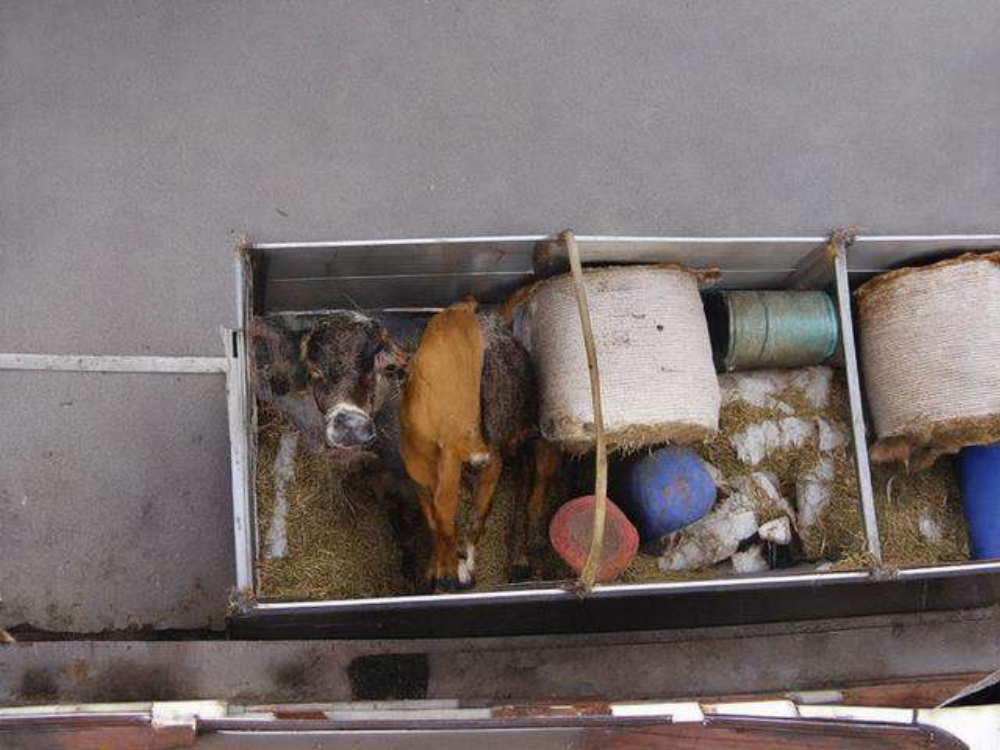}
        \caption*{\footnotesize (h) Inverted (Hard to invert)}
    \end{subfigure}

    \caption{\textbf{Examples of images and their reconstructions using our inversion method.} Top two rows: easy-to-invert images with high reconstruction fidelity. Bottom two rows: hard-to-invert images with noticeable semantic and structural degradation.}
    \label{fig:inversion_examples}
\end{figure*}

Figure~\ref{fig:inversion_examples} displays four such images along with their reconstructions using our inversion method. The top two rows correspond to easy-to-invert samples, where the inverted outputs closely match the originals in both low-level and semantic structure. The bottom two rows show hard-to-invert cases, where the reconstructions diverge significantly in content or structure, despite the input being real. These qualitative results support our core claim that not all real images can be reliably certified post-hoc, particularly when inversion fidelity is low.

\begin{figure*}[!t]
    \centering
    \includegraphics[width=\textwidth]{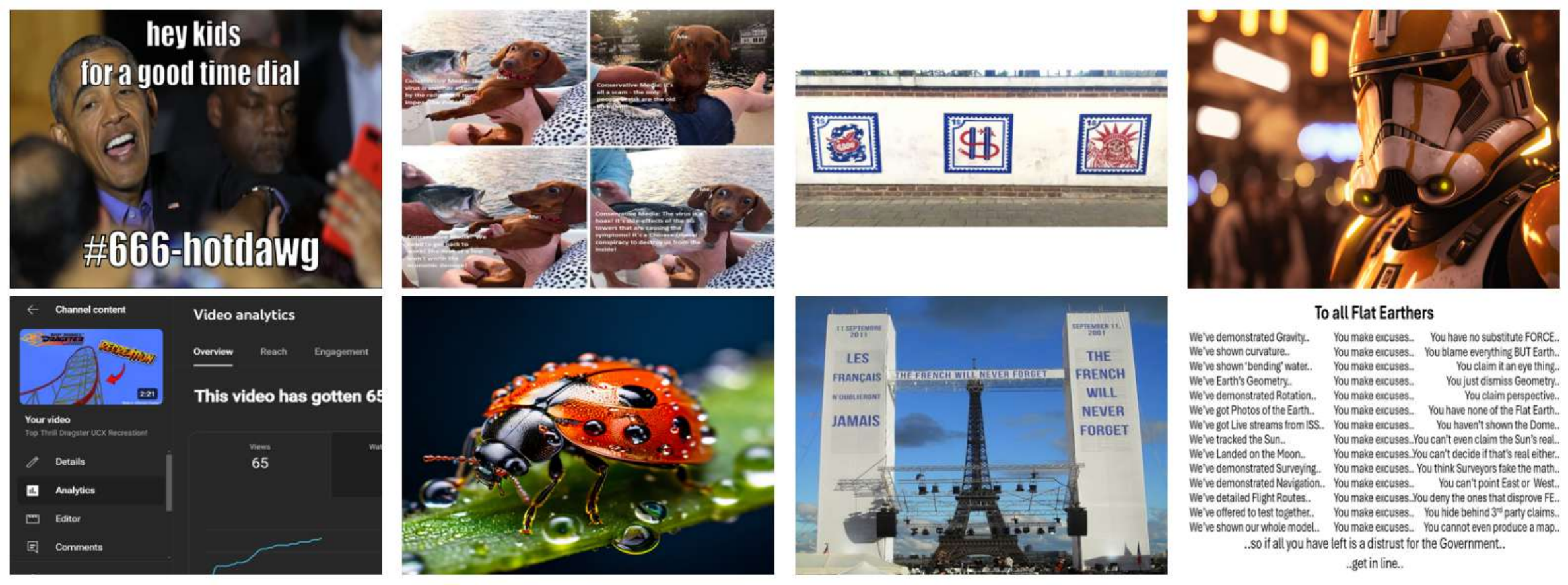}
    \caption{Collage of 8 real images scraped from the internet. Despite their authenticity, most fall below the A-index safety threshold, indicating plausible deniability due to high-fidelity inversion.}
    \label{fig:scraped_collage_20}
\end{figure*}

\section{Scraped Internet Images: Hard-to-Invert and Easy-to-Invert Examples}

To assess the real-world implications of our method, we analyze a corpus of approximately 3,000 images scraped from Reddit across diverse topics. Each image is inverted using Stable Diffusion 3 (medium) and scored using the A-index. Based on the calibrated safety threshold for this model, we categorize images as either likely authentic (above threshold) or plausibly deniable (below threshold).

\begin{figure*}[t]
    \centering

    \begin{subfigure}[b]{0.23\linewidth}
        \includegraphics[width=\linewidth]{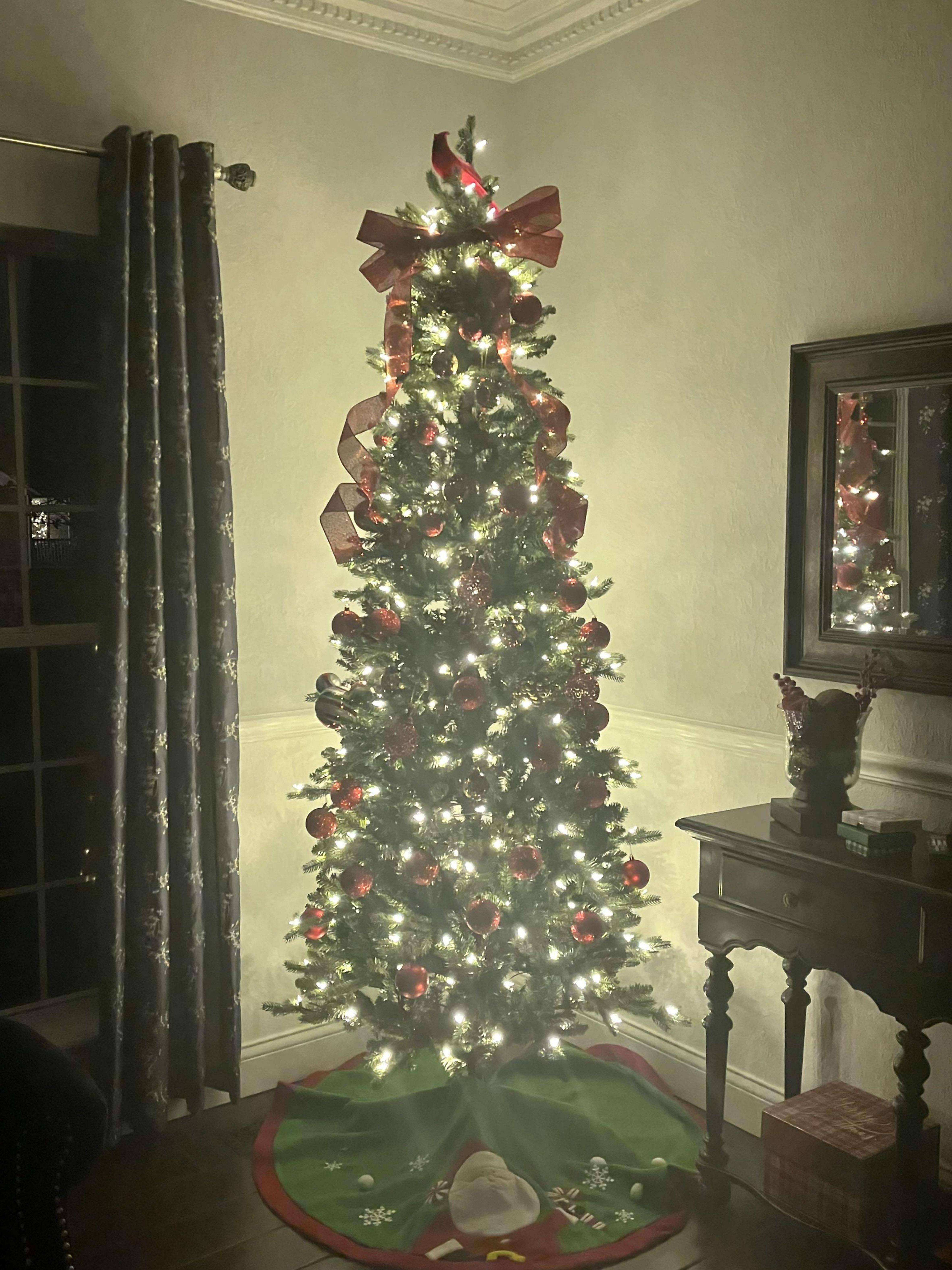}
        \caption{Original (Easy to invert)}
    \end{subfigure}
    \hfill
    \begin{subfigure}[b]{0.23\linewidth}
        \includegraphics[width=\linewidth]{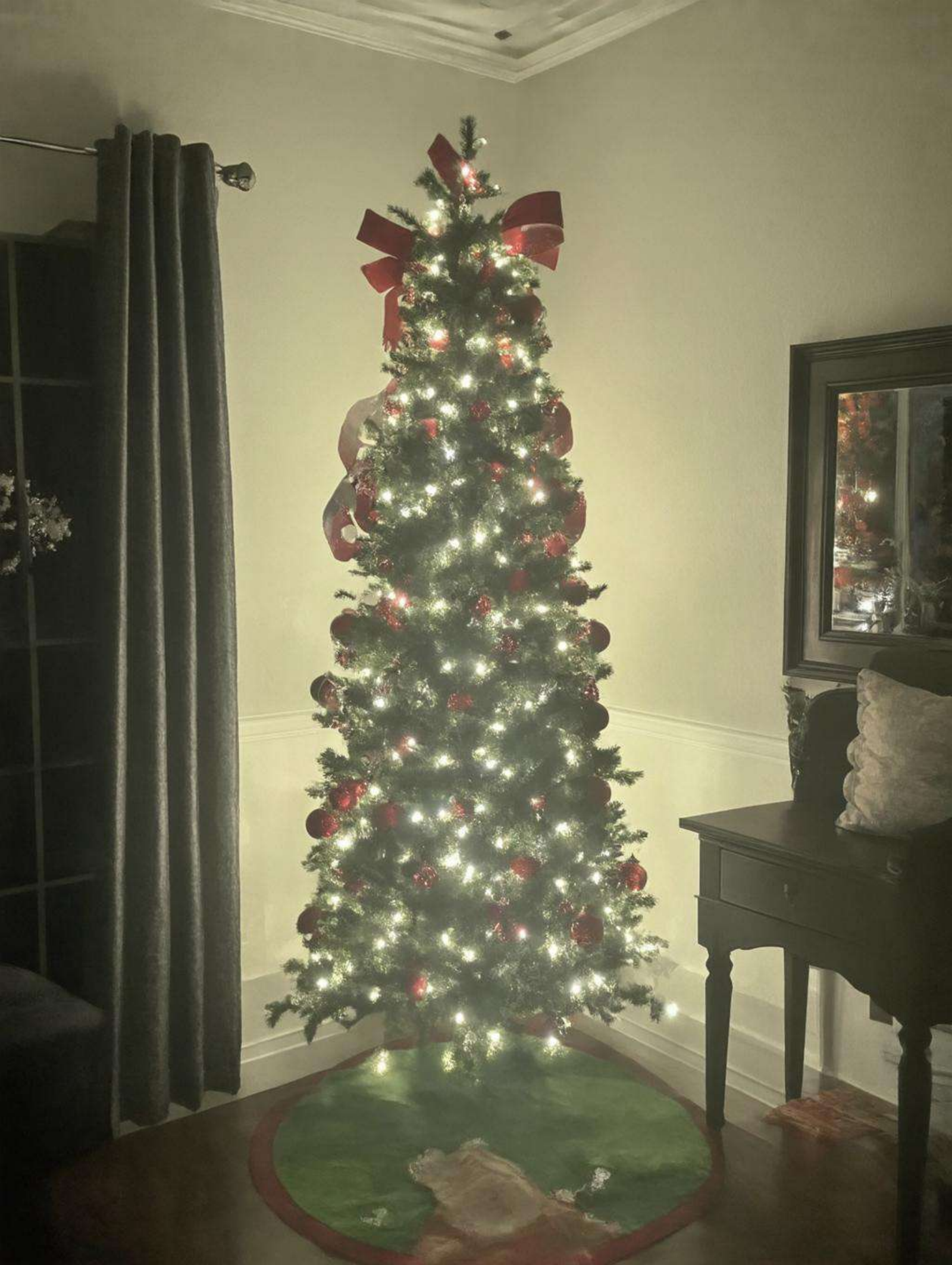}
        \caption{Inverted (Easy to invert)}
    \end{subfigure}
    \hfill
    \begin{subfigure}[b]{0.23\linewidth}
        \includegraphics[width=\linewidth]{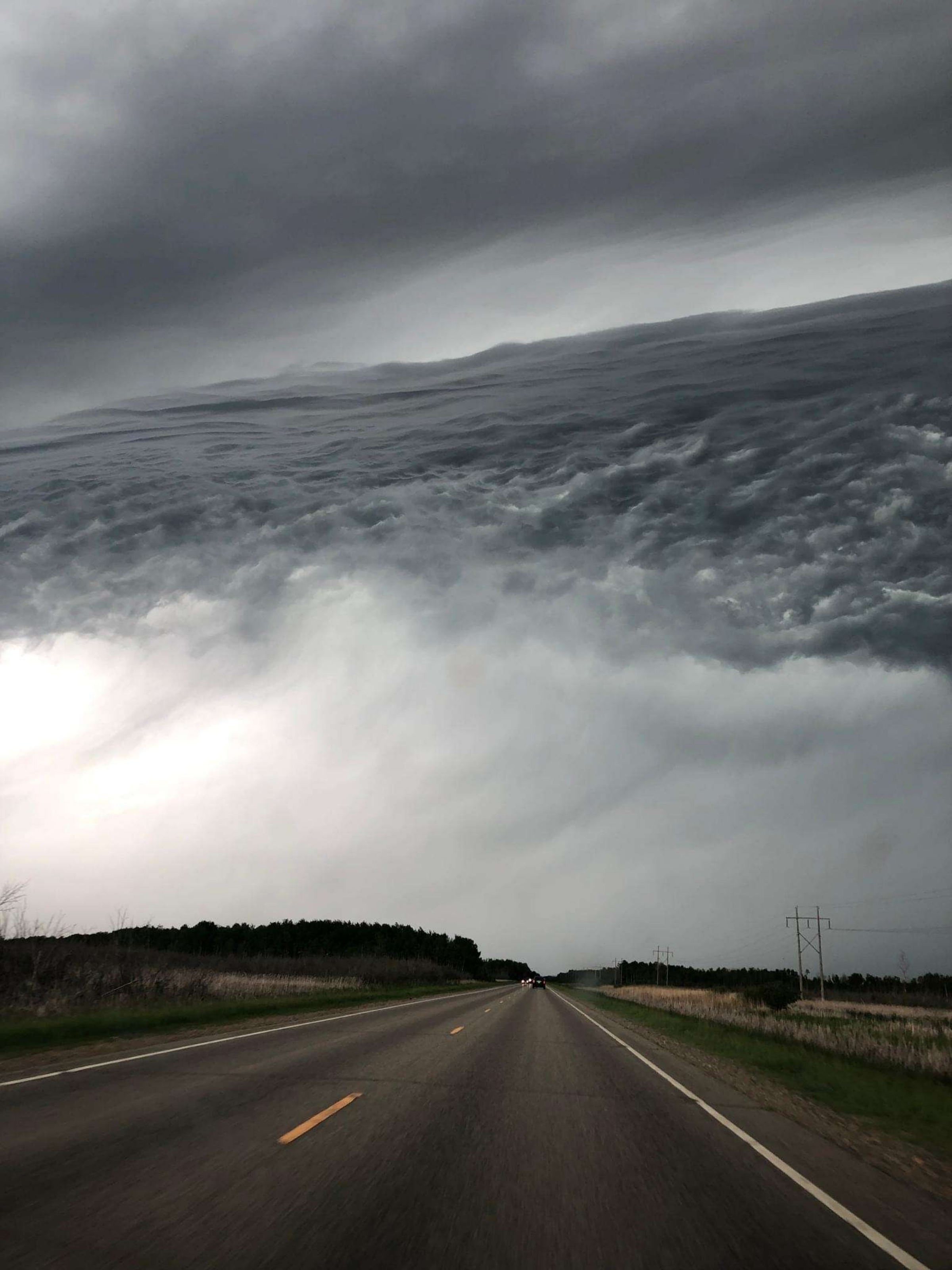}
        \caption{Original (Easy to invert)}
    \end{subfigure}
    \hfill
    \begin{subfigure}[b]{0.23\linewidth}
        \includegraphics[width=\linewidth]{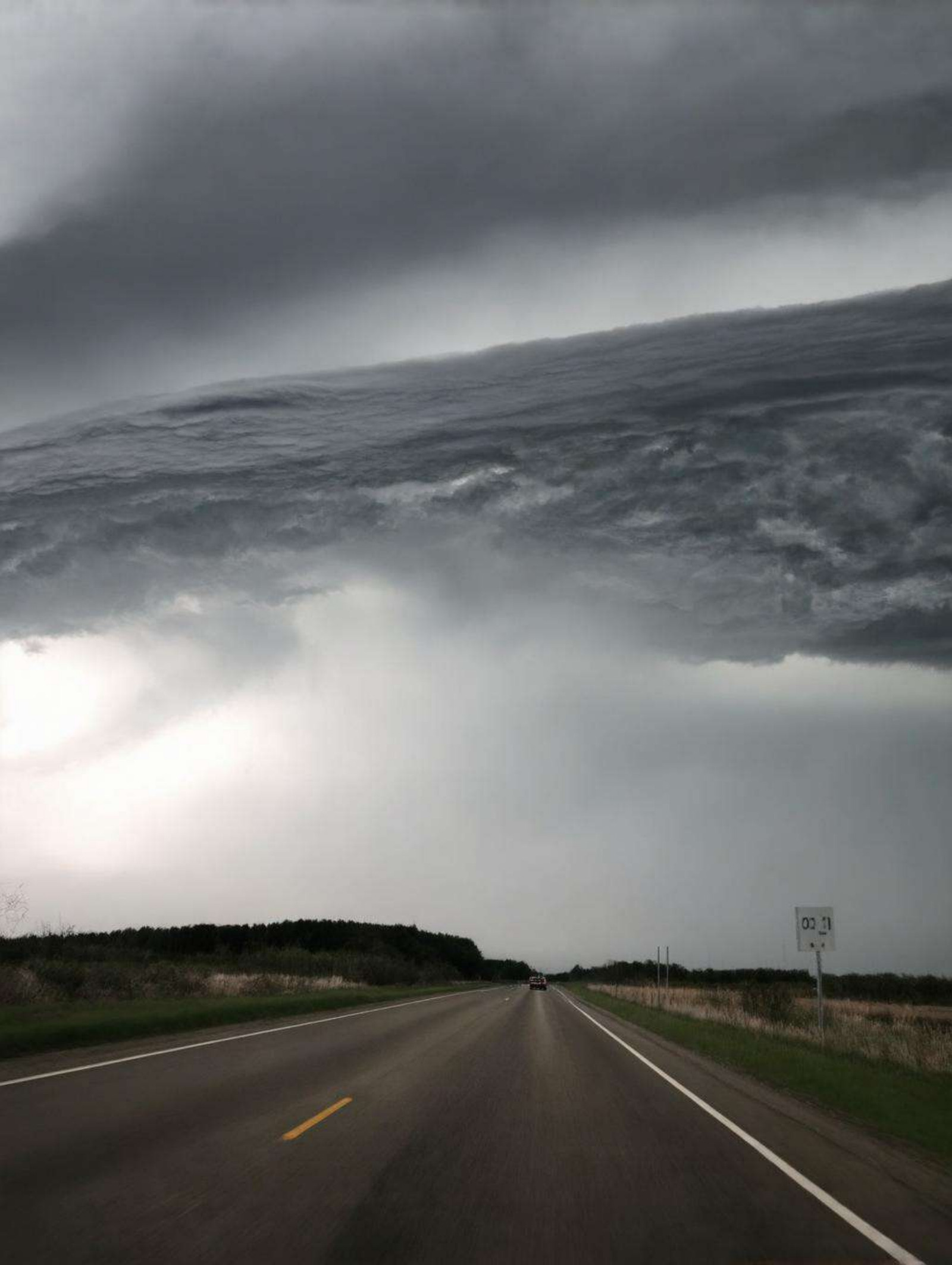}
        \caption{Inverted (Easy to invert)}
    \end{subfigure}

    \vspace{0.4cm}

    \begin{subfigure}[b]{0.23\linewidth}
        \includegraphics[width=\linewidth]{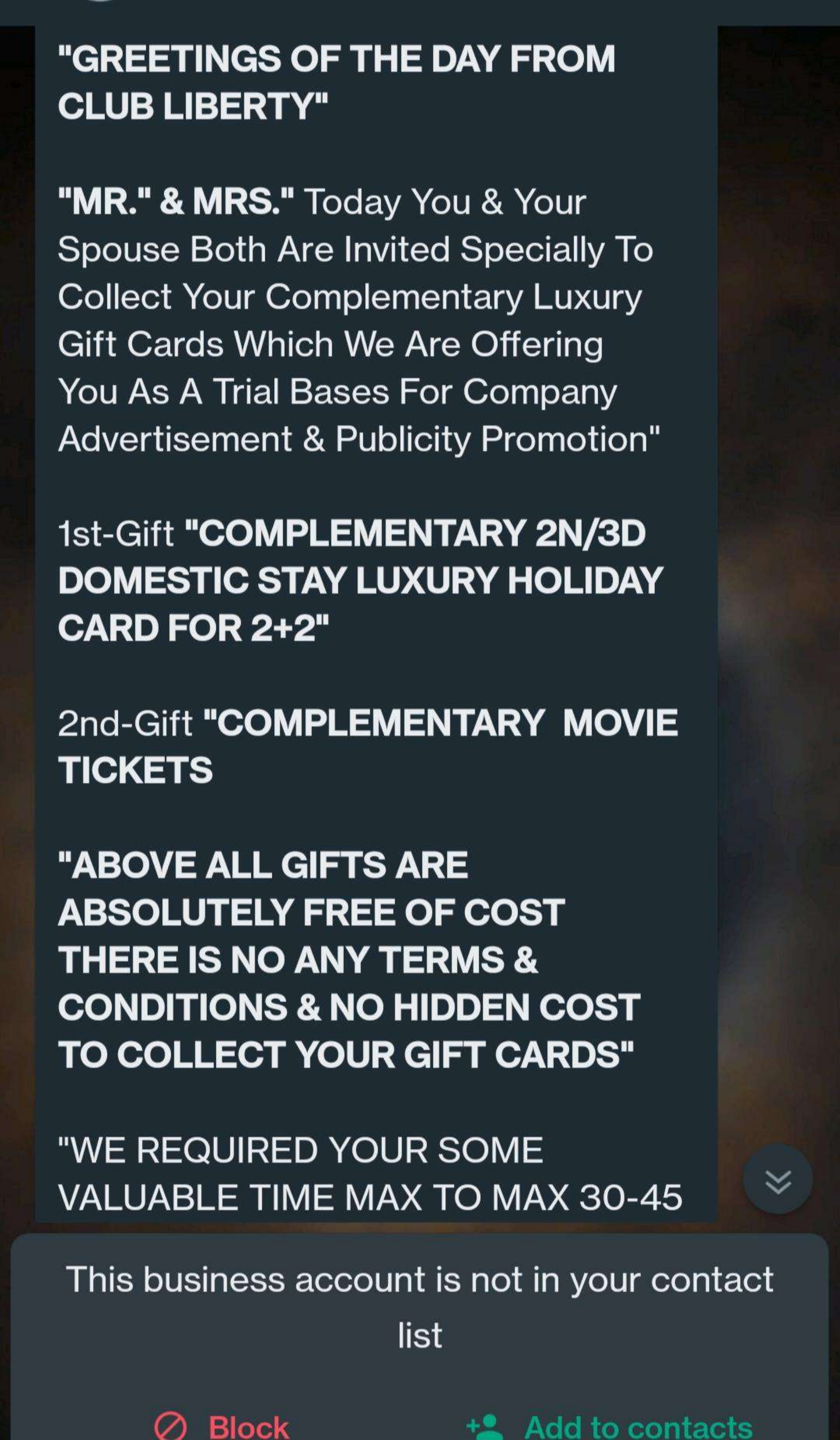}
        \caption{Original (Hard to invert)}
    \end{subfigure}
    \hfill
    \begin{subfigure}[b]{0.23\linewidth}
        \includegraphics[width=\linewidth]{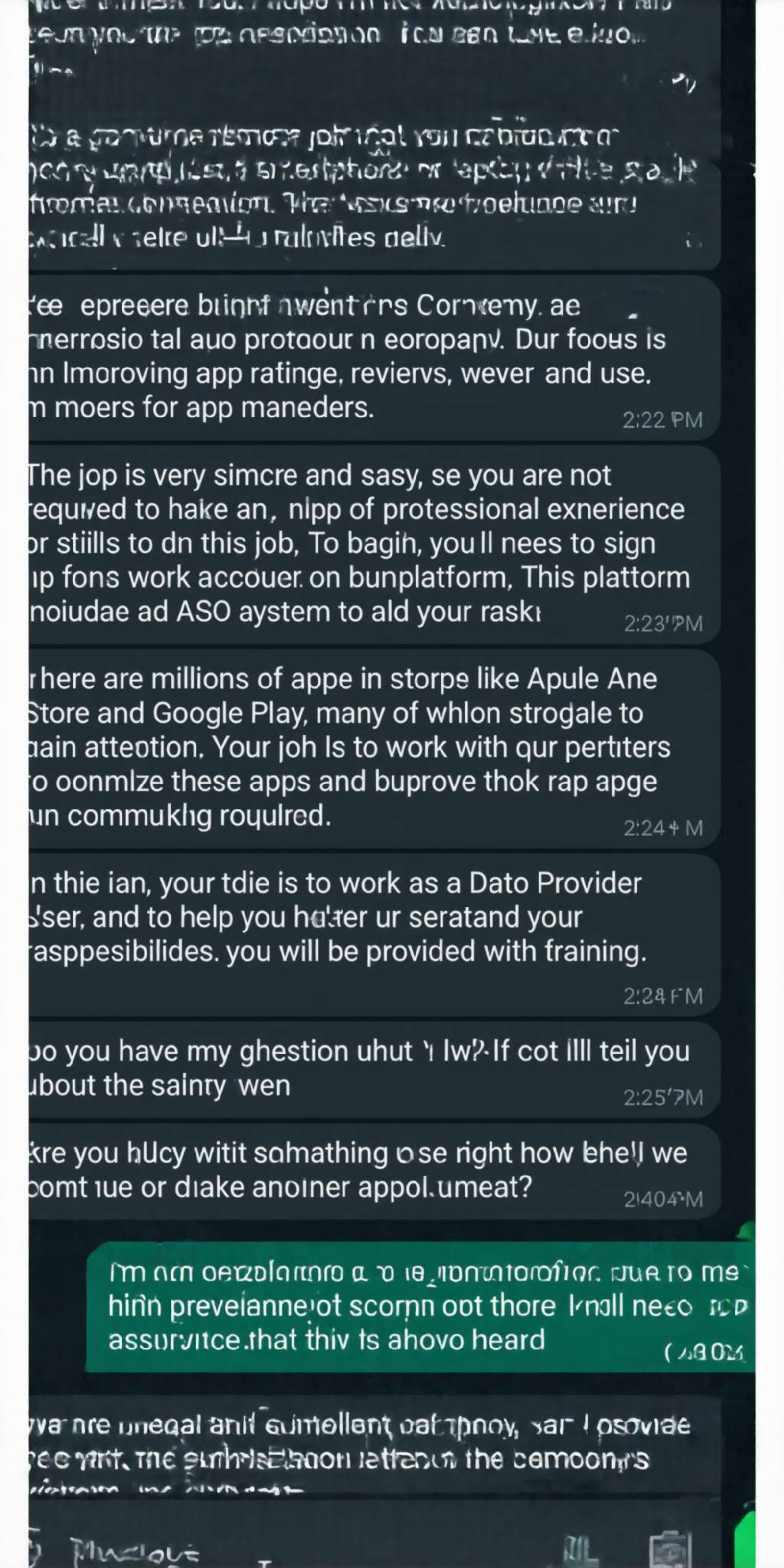}
        \caption{Inverted (Hard to invert)}
    \end{subfigure}
    \hfill
    \begin{subfigure}[b]{0.23\linewidth}
        \includegraphics[width=\linewidth]{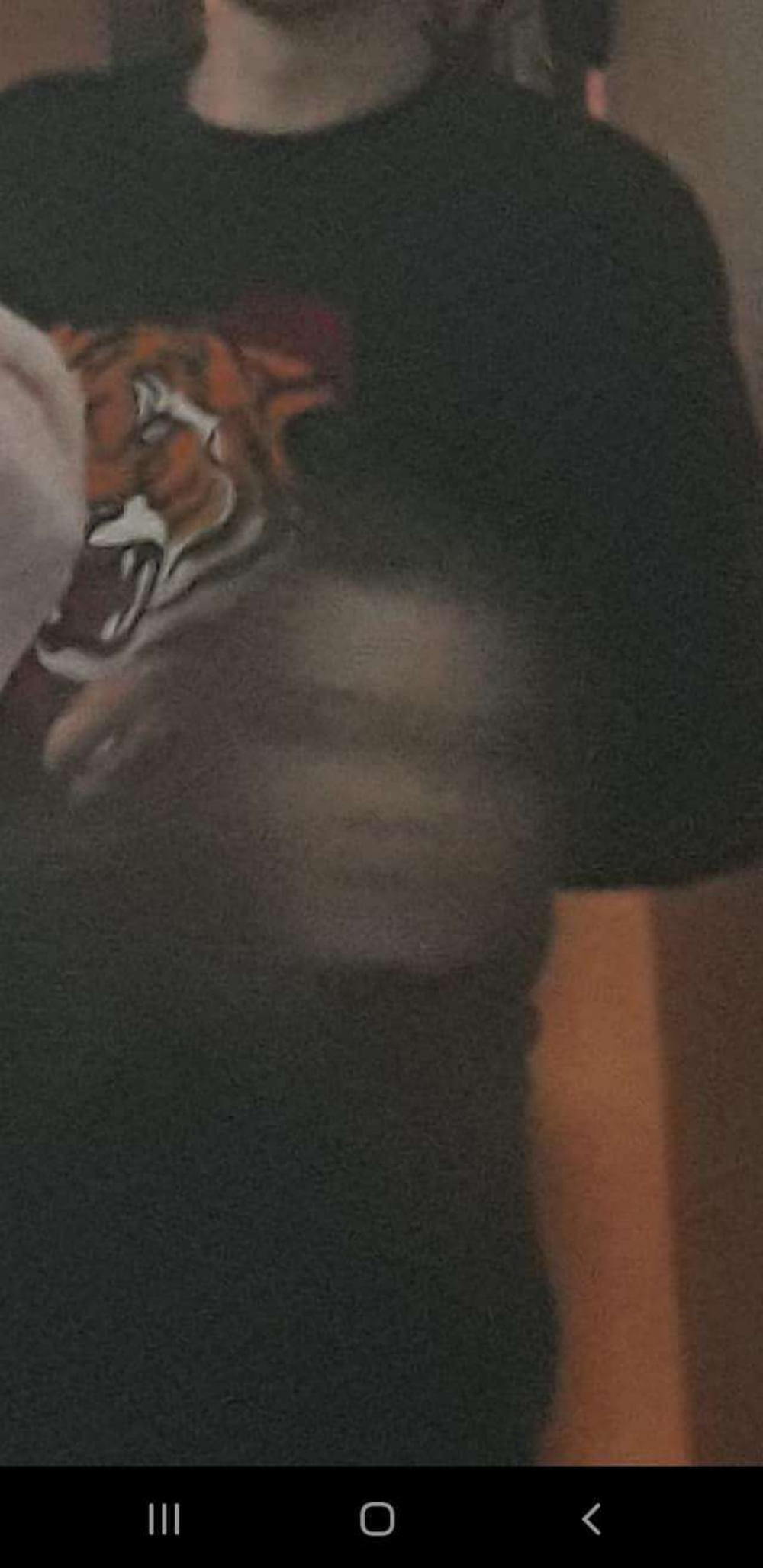}
        \caption{Original (Hard to invert)}
    \end{subfigure}
    \hfill
    \begin{subfigure}[b]{0.23\linewidth}
        \includegraphics[width=\linewidth]{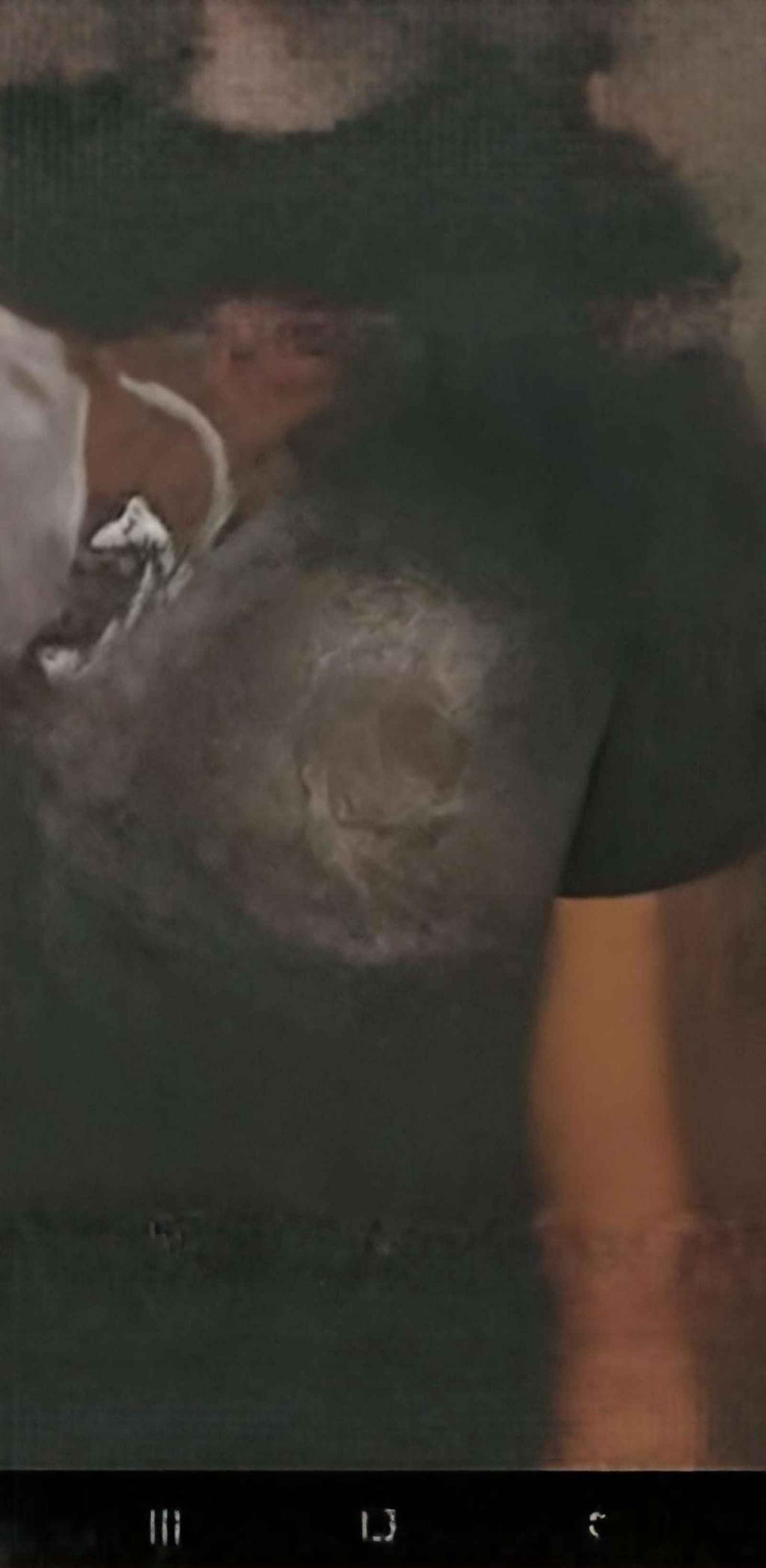}
        \caption{Inverted (Hard to invert)}
    \end{subfigure}

    \caption{Comparison of inversion performance on scraped internet images. The top row displays two examples that produce accurate, high-quality reconstructions. The bottom row shows cases where the inversion process struggles, often due to fine text, low contrast, or complex object structure, resulting in substantially degraded reconstructions.}
    \label{fig:scraped_examples}
\end{figure*}

Figure~\ref{fig:scraped_examples} shows two representative cases: one image that is easy to invert, and one that is hard to invert. The easy-to-invert image reconstructs with high perceptual and semantic fidelity, resulting in a low A-index score. In contrast, the hard-to-invert image exhibits substantial degradation in the inverted output, despite being real. These examples further demonstrate that some authentic content cannot be reliably verified post-hoc, especially when generative models closely match the visual features.

To understand how this manifests across broader internet content, we include a visual summary in Figure~\ref{fig:scraped_collage_20}, which displays eight scraped images from different Reddit communities. Despite all being real, only a handful exceed the safety threshold, indicating how widespread plausible deniability has become in online imagery.

\section{Medium Resource Attacker Full Details}
\label{sec:medium-attacker-full-details}

To operationalize our medium-resource attacker scenario , we simulate a threat model where the adversary is granted access to a single prompt and a fixed computational budget. The attacker samples \(N = 100\) images by drawing random seeds and generating samples \(x_i = G_\theta(z_i;\text{prompt})\) for \(i = 1,\dots,100\), where \(G_\theta\) is the target generative model (Stable Diffusion 3 medium in our case). Each generated image is inverted using our reconstruction-free inversion pipeline \(\widetilde{G}^{-1}\) and scored using the Authenticity Index:
\[
\text{A-index}(x_i, \widetilde{x}_i) = \text{A-index}(x_i, \widetilde{G}^{-1}(x_i)).
\]

The attacker then selects the highest-scoring candidate \(x^\star\) and optimizes a small perturbation \(\delta\), constrained by \(\|\delta\|_\infty \leq 8/255\), to maximize the A-index after inversion of the perturbed image.

The prompt used in this experiment is:
\begin{quote}
\myverbatim{A hungry man standing outside a real pizza shop at night, mouth slightly open, drooling, pointing toward the glowing neon pizza sign. Warm light from the shop window reveals fresh cheesy pizzas inside with steam, the scene looks realistic and cinematic with natural lighting and lifelike details.}
\label{quote:prompt_hungry_man_real_pizza}
\end{quote}

To illustrate the range of outputs available under the attacker's sampling budget, we visualize six randomly sampled images generated from this prompt using different seeds in Figure~\ref{fig:medium_attacker_variants}. These images are unmodified generations prior to inversion or adversarial refinement. While visually diverse, the top-scoring image among the 100 candidates achieved an \(\text{A-index}\) of 0.0148. After applying PGD to perturb \(x^\star\), the \(\text{A-index}(x^\star, \widetilde{G}^{-1}(x^\star + \delta))\) increased modestly to 0.0154. This slight improvement remains below both the safety threshold (\(\tau_\text{safety} = 0.0365\)) and the adversarial security threshold (\(\tau_\text{security} = 0.038\)), indicating the limited effectiveness of a medium-resource attacker.

\section{Limitations}

Despite its strengths, our approach has several limitations that highlight directions for future research.

First, the authenticity index relies on access to high-quality inversion pipelines and perceptual similarity models such as CLIP and LPIPS. While this enables a robust and calibrated score, it assumes white-box or partially open generative models. In scenarios where the generator is fully black-box or proprietary, the inversion step may not be feasible or reliable.

second, the safety and security thresholds are specific to each generative model and require calibration using real and synthetic data. This introduces practical overhead, especially in dynamic environments where new models are rapidly emerging. Automating or generalizing this calibration remains an open challenge.

Finally, while our extension to the video domain is promising, it treats each frame independently and does not exploit temporal consistency or motion cues. Incorporating these temporal dynamics could improve performance in challenging video-based deepfake scenarios.

\end{document}